\documentclass[10pt,twocolumn,letterpaper]{article}
\usepackage[toc,page]{appendix}
\makeatletter
\@namedef{ver@everyshi.sty}{}
\makeatother
\usepackage{titling}
\usepackage[pagenumbers]{cvpr} 
\usepackage{graphicx}
\usepackage[T1]{fontenc}
\usepackage{amsmath}
\usepackage{amssymb}
\usepackage{float}
\usepackage{booktabs}
\usepackage{tablefootnote}
\usepackage{longtable}
\usepackage{tabularx}
\usepackage{multirow}
\usepackage{boldline}
\usepackage{verbatim}
\usepackage{graphicx}
\usepackage{pifont}
\usepackage{caption,subcaption,tikz}
\usepackage{etoolbox}
\BeforeBeginEnvironment{appendices}{\clearpage}
\usepackage[pagebackref,breaklinks,colorlinks]{hyperref}
\usepackage[capitalize]{cleveref}
\crefname{section}{Sec.}{Secs.}
\Crefname{section}{Section}{Sections}
\Crefname{table}{Table}{Tables}
\crefname{table}{Tab.}{Tabs.}

\newcommand{\expect}[1]{\ensuremath{\operatorname{\mathbb{E}}\!\left[ #1 \right]}}
\newcommand{\var}[1]{\ensuremath{\operatorname{Var}\!\left[ #1 \right]}}
\begin{document}

\title{The Norm Must Go On: Dynamic Unsupervised Domain Adaptation by Normalization}
\author{M. Jehanzeb Mirza$^{1,2}$
\and
Jakub Micorek$^{1}$
\and
Horst Possegger$^{1}$
\and
Horst Bischof$^{1,2}$\\
\and
$^{1}$\text{Institute for Computer Graphics and Vision, Graz University of Technology.}\\ 
$^{2}$\text{Christian Doppler Laboratory for Embedded Machine Learning.}\\ 
{\tt\small \{muhammad.mirza, jakub.micorek, possegger, bischof\}@icg.tugraz.at}
}
\maketitle

\begin{abstract}
Domain adaptation is crucial to adapt a learned model to new scenarios, such as domain shifts or changing data distributions. Current approaches usually require a large amount of labeled or unlabeled data from the shifted domain. This can be a hurdle in fields which require continuous dynamic adaptation or suffer from scarcity of data, \eg autonomous driving in challenging weather conditions. To address this problem of continuous adaptation to distribution shifts, we propose \underline{D}ynamic \underline{U}nsupervised \underline{A}daptation (DUA). 
By continuously adapting the statistics of the batch normalization layers we modify the feature representations of the model.
We show that by sequentially adapting a model with only a fraction of unlabeled data, a strong performance gain can be achieved. 
With even less than 1\% of \emph{unlabeled data} from the target domain, DUA already achieves competitive results to strong baselines. In addition, the computational overhead is minimal in contrast to previous approaches. Our approach is simple, yet effective and can be applied to any architecture which uses batch normalization. We show the utility of DUA by evaluating it on a variety of domain adaptation datasets and tasks including object recognition, digit recognition and~object detection.
\end{abstract}
\section{Introduction}
\label{sec:intro}
Present day Deep Neural Networks (DNNs) show promising results when both training and testing data belong to the same distribution~\cite{krizhevsky2012imagenet, he2016deep, xie2017aggregated}. However, if there is a domain shift, \ie when the testing data comes from a different domain, neural networks struggle to generalize~\cite{luo2019taking, chen2018domain, ganin2016domain}. In fact, even if there is only a slight distribution shift, the performance of neural networks is reported to already degrade significantly~\cite{hendrycks2019robustness, recht2019imagenet}.

\begin{figure}
    \centering
    
    \begin{subfigure}[b]{0.47\textwidth}
            
            \centering
            \includegraphics[width=8cm,height=6cm]{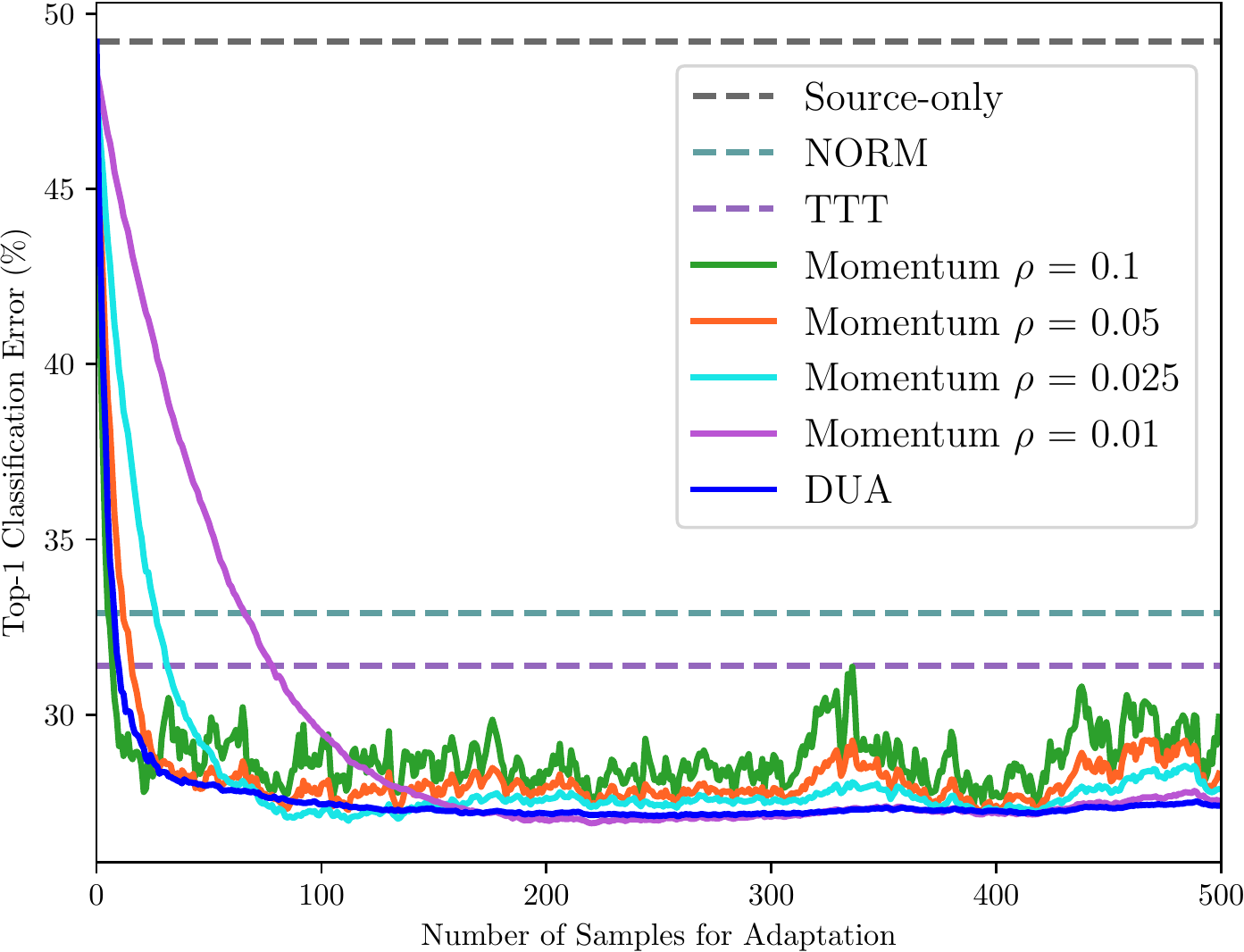}
            \caption{CIFAR-10C results.}
            \label{fig:comparison_bn_mom}
    \end{subfigure}%
    
   \begin{subfigure}[b]{0.47\textwidth}
            \centering
            \includegraphics[scale = 0.53]{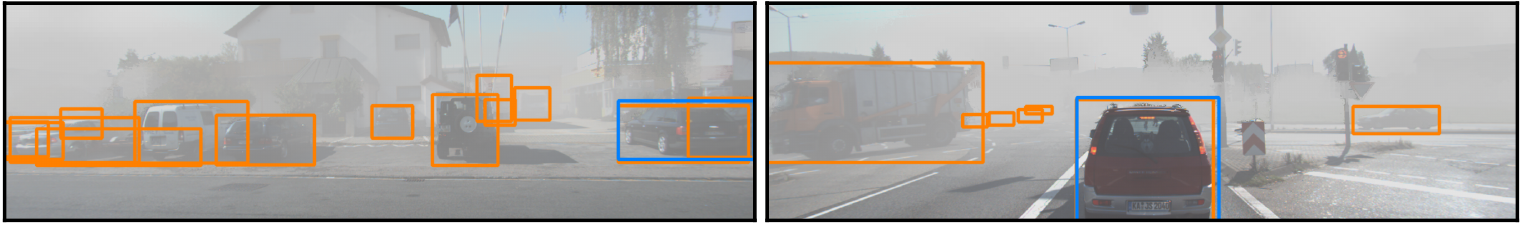}
            \caption*{}
            \label{fig:detection_visuals_unadapted}
    \end{subfigure}%
    
    \vspace{-3mm}\begin{subfigure}[b]{0.47\textwidth}
            \centering
            \includegraphics[scale = 0.53]{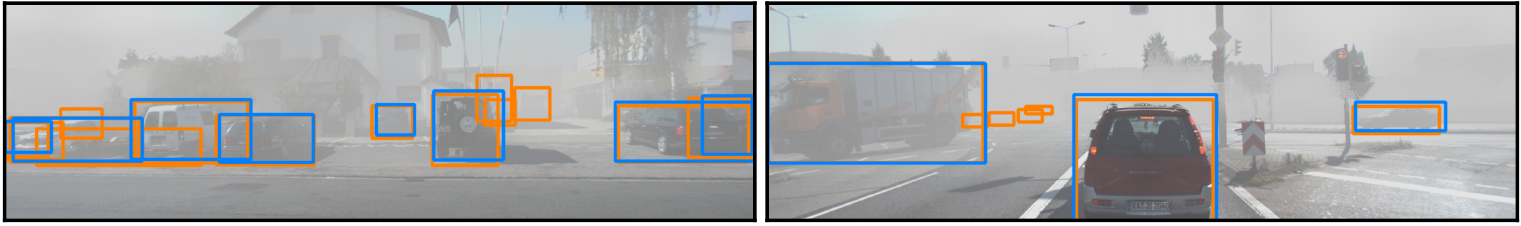} 
            \caption{Detections in foggy weather without (top) and with (bottom) DUA. We overlay the detection results (blue) and the ground truth (orange). }
            \label{fig:detection_visuals_adapted}
    \end{subfigure}%
    \caption{Exemplary DUA results. a) Mean classification error over 15 different corruption types (at the most severe level 5) on CIFAR-10C~\cite{hendrycks2019robustness}. We outperform the state-of-the-art NORM~\cite{schneider2020improving, nado2020evaluating} and TTT~\cite{sun2020test}, while using less than 1\% of unlabeled data from the corrupted test set only. Our proposed adaptive momentum scheme leads to both fast and stable improvements in contrast to fixing the momentum parameter $\rho$.
    b) Qualitative results for object detection in degrading weather conditions: our DUA (bottom) significantly improves the performance of a KITTI~\cite{geiger2013vision} pre-trained YOLOv3~\cite{redmon2018yolov3} on KITTI-Fog~\cite{halder2019physics}. Best viewed in color.}
    \label{fig:comparison}
\end{figure}

One way to overcome the performance drop during domain shifts is to obtain labeled data from the shifted domain and re-train the network. However, manual labeling of large amounts of data imposes significant human and monetary costs.
These issues are addressed by Unsupervised Domain Adaptation (UDA) approaches, \eg~\cite{chen2018domain,gidaris2018unsupervised,ganin2016domain, liu2019transferable, xie2019multi, zheng2020cross, he2019multi, saito2019strong, kim2019diversify, liu2016coupled}.
For UDA, the goal is to modify the network parameters in such a way that it can adapt in an unsupervised manner to out-of-distribution testing data. Traditionally, these approaches require labeled training data along with a large amount of unlabeled testing data.  

In many practical scenarios, the traditional requirements, \ie access to both labeled training and large amounts of unlabeled testing data can often not be fulfilled. For example, in the medical domain, pre-trained models are often provided without access to the training data (which is kept private due to privacy regulations). Likewise, some application domains benefit from dynamic adaptation to a changing environment.
For example, consider object detectors for autonomous vehicles, which are usually trained on mostly clear weather images, \eg~\cite{geiger2013vision, han2021soda10m, sun2020scalability}.
In real-world scenarios, however, weather can suddenly deteriorate, resulting in significant performance degradation~\cite{mirza2021robustness, michaelis2019benchmarking}. In such cases, it is not feasible to obtain labeled training data captured in degrading weather and re-train the detector from scratch. A better solution is to dynamically adapt the detector, given only a few (unlabeled) bad weather examples.

In this work, we show that one hindrance in domain generalization is the statistical difference in mean and variance between train and (shifted) test data. Thus, during inference, we adapt the running mean and variance which are calculated during training by the batch normalization layer~\cite{ioffe2015batch}. Moreover, we adapt the statistics dynamically in an online manner on a tiny fraction of test data. For adaptation, we form a small batch by augmenting each incoming sample. In order to ensure stable adaptation and fast convergence we propose an adaptive update schema.  

Related approaches~\cite{li2016revisiting, schneider2020improving, nado2020evaluating, wang2020tent} typically ignore the training statistics and recalculate the batch statistics from scratch for the test data. This, however, requires large batches of test data. We argue that a large batch of test data might not always be available in real-world applications, \eg autonomous cars adapting to challenging weather (see Figure~\ref{fig:comparison}). We show that a strong performance gain can be achieved by adapting the running mean and variance in an online manner (one sample at a time). In particular, we require only a small number of sequential samples from the out-of-distribution data. 

Our contributions can be summarized as follows:

\begin{itemize}
    \item We show that online adaptation of batch normalization parameters on a tiny fraction of unlabeled out-of-distribution test data can provide a strong performance gain. With even less than 1\% of unlabeled test data, DUA already performs competitively to strong baselines which use the entire test set for adaptation.  
    \item DUA is simple, unsupervised, dynamic and requires no back propagation~\cite{rumelhart1986learning} to work. Since the computational overhead is also negligible, it is perfectly suited for real-time applications.   
    \item We evaluate DUA on a variety of domain shift benchmarks, demonstrating its strong performance. We achieve state-of-the-art results on most benchmarks while being competitive on the remaining.  
    \item We show that our dynamic adaptation method works on a variety of different tasks and different architectures. To the best of our knowledge, we are the first to show dynamic adaptation for object detection.  
\end{itemize}
\section{Related Work}
\label{sec:related_work}
Unsupervised Domain Adaptation (UDA) has received a significant amount of interest recently. We summarize these approaches in four categories: minimizing discrepancy between domains, adversarial approaches, self-supervised approaches and correcting domain statistics.

\textbf{Discrepancy reduction} between source and target domains is usually performed at specific network layers or in a contrastive manner. Long et al.~\cite{long2015learning} match the mean embeddings from task specific layers. Sun et al.~\cite{sun2017correlation, sun2016deep} minimize the second order statistics to align the source and target domains. They apply a linear transformation on the source domain to align it with the target domain. Zellinger et al.~\cite{zellinger2017central} propose to match higher order moments by introducing a Central Moment Discrepancy (CMD) metric to learn domain invariant features. Chen et al.~\cite{chen2020homm} propose to match third and fourth order statistics of the source and target domains for unsupervised domain adaptation. On the other hand,~\cite{wang2021cross, kang2019contrastive} use contrastive learning~\cite{chopra2005learning} to reduce discrepancy between domains.

 \textbf{Adversarial discriminative} approaches align the features from the source and target domains mostly by using the domain confusion loss. Ganin et al.~\cite{ganin2016domain} propose a method which is based on the philosophy that predictions must be made on features which are non-discriminative during training. A novel gradient reversal layer is proposed which brings the features from the source and target domains closer by maximizing the domain confusion loss. Tzeng et al.~\cite{tzeng2017adversarial} also rely on maximizing the domain confusion loss for unsupervised domain adaptation. Hong et al.~\cite{hong2018conditional} use a fully convolutional network and use generative adversarial networks~\cite{goodfellow2014generative} to address the problem of synthetic-to-real feature alignment. Chen et al.~\cite{chen2017no} align the global and class wise features by using a generative adversarial network. Similar approaches for UDA have also been followed for a variety of tasks, including object detection~\cite{he2019multi,chen2018domain,kim2019diversify,wang2020train, xie2019multi,xu2021spg,yang2021st3d,zheng2020cross,fruhwirth2021fast3d}, object classification~\cite{liu2019transferable,liu2016coupled,liu2021recursively,qin2019pointdan,rebut-shot} and semantic segmentation~\cite{jaritz2020xmuda,chen2021unsupervised,zou2018unsupervised,biasetton2019unsupervised,yu2021dast, lian2019constructing} for both 2D and 3D data. 

\textbf{Self Supervision}
has also been used for the purpose of unsupervised domain adaptation. Sun et al.~\cite{sun2019unsupervised} combine different self supervised auxiliary tasks for domain adaptation. Sun et al.~\cite{sun2020test} also propose Test Time Training (TTT) with self supervision. They put forward the idea of removing the self-imposed condition of a fixed decision boundary at test time. In their work they use the rotation prediction task~\cite{gidaris2018unsupervised} as a self supervised task in order to adapt the network to out-of-distribution test data.

\textbf{Correcting the domain statistics} calculated by the batch normalization layer~\cite{ioffe2015batch} has also been used for UDA. Li et al.~\cite{li2016revisiting} propose Adaptive Batch Normalization where they show that recalculating the batch normalization parameters from scratch for the test set can improve generalization of DNNs. Carlucci et al.~\cite{maria2017autodial} propose domain adaptation layers which can learn a hyperparameter during training to find the optimal mixing of statistics from the source and target domain. Singh et al.~\cite{singh2019evalnorm} study the effect of lower batch sizes during training and show that DNNs using batch normalization layers are effected by lower batch sizes. They propose an auxiliary loss for remedy. Similarly,~\cite{nado2020evaluating, schneider2020improving, arm} also show that recalculation of batch normalization statistics from scratch for test data can be helpful to address the problem of distribution shift between source and target domains. Wang et al.~\cite{wang2020tent} also recalculate the batch normalization statistics for the test data. Further, they calculate the loss from the entropy of predictions and adapt the scale and shift parameters of batch normalization layers. It is important to point out that~\cite{li2016revisiting, nado2020evaluating, schneider2020improving, wang2020tent} share the same philosophy of a variable decision boundary at test time as TTT~\cite{sun2020test}. 

Our work closely resonates with~\cite{li2016revisiting, nado2020evaluating, schneider2020improving, wang2020tent} and is also similar in philosophy to TTT~\cite{sun2020test}, aiming for a variable decision boundary at test time. However, we differ from them in several fundamental ways: In~\cite{li2016revisiting, nado2020evaluating, schneider2020improving, wang2020tent}, the training statistics are ignored and the batch statistics are recalculated from the test set. For this reason they require large batches from the test set. However, in general, large batches of test data might not be available. Instead, we adapt the statistics calculated from the training data in an online manner (on each incoming sample). We show competitive results by using less than 1\% of unlabeled test data in contrast to all previous approaches which use the complete test set. Further, contrary to previous approaches, such as~\cite{wang2020tent,sun2020test}, our method does not require back propagation. Our scenario is more realistic for dynamic adaptation where we can obtain only a single test frame at one time.    
\begin{figure*}
\centering
\begin{subfigure}{.5\textwidth}
  \centering
  \includegraphics[scale = 0.78]{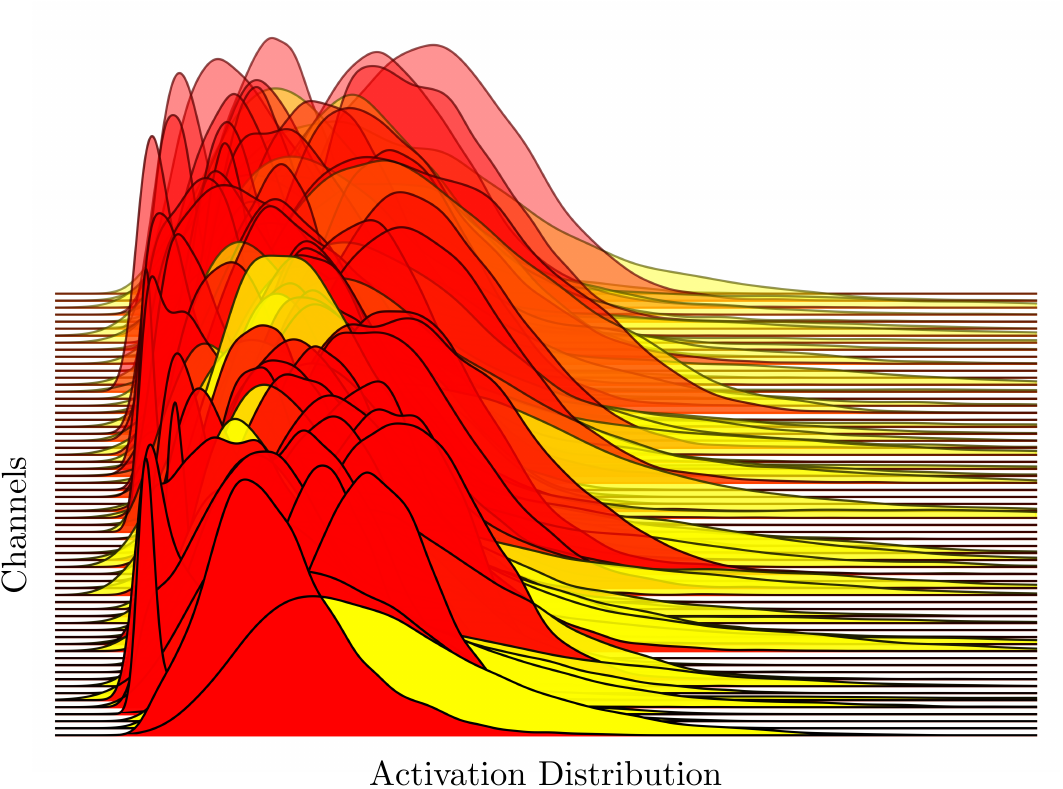}
  \caption{Source Model}
  \label{fig:sub1}
\end{subfigure}%
\begin{subfigure}{.5\textwidth}
  \centering
  \includegraphics[scale = 0.78]{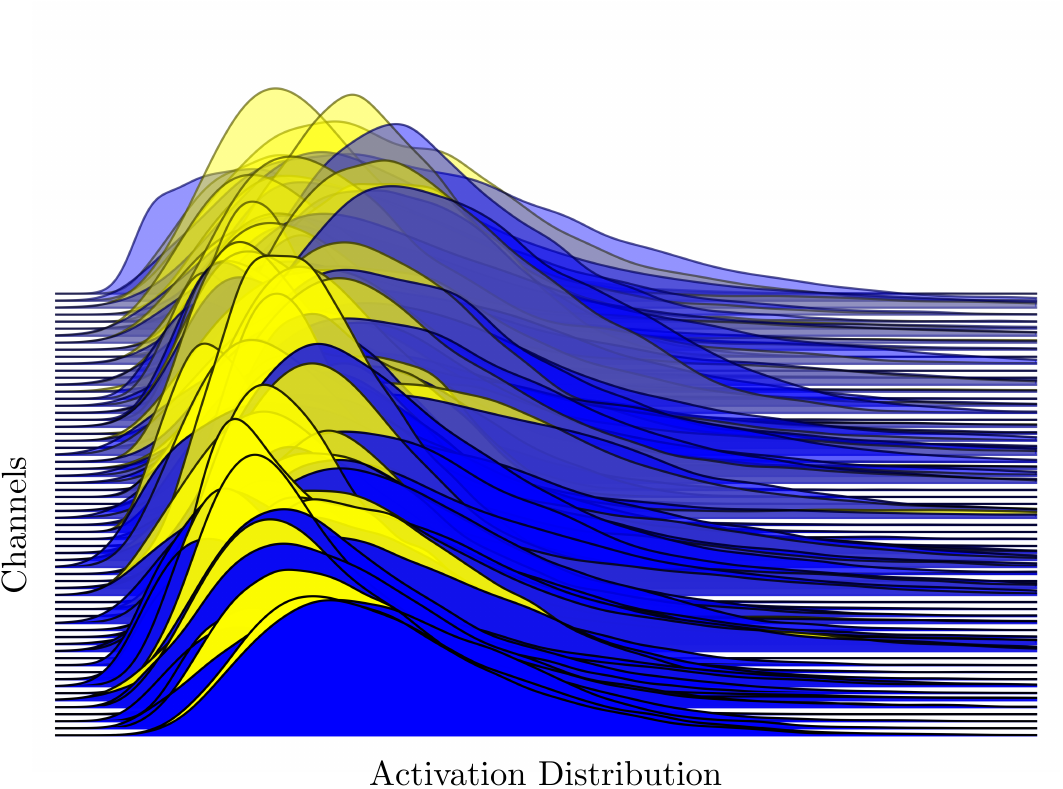}
  \caption{Model Adapted with DUA}
  \label{fig:sub2}
\end{subfigure}
\caption{Density plots of output distribution for the 64 channels of the last batch normalization layer of ResNet-26 trained on CIFAR-10. a) Yellow is the output distribution of the training data. Red is the output distribution of the shifted test data, \ie corrupted with Contrast Level-5~\cite{hendrycks2019robustness}. The misalignment of the feature responses is one reason for the performance drop. b) Yellow is the output distribution of the training data. Blue is the output distribution of the shifted test data, \ie corrupted with Contrast Level-5, after adaptation with DUA. DUA aligns the output distribution from corrupted data closely with the clean (training) distribution. Best viewed in color.}
\label{fig:feature_visualization}
\end{figure*}

\section{Approach}
\label{sec:approach}

We first summarize batch normalization \cite{ioffe2015batch} in Section~\ref{subsec:batch_normalization_description} as it lies at the center of our approach. Section~\ref{subsec:method} then details our DUA approach.

\subsection{Batch Normalization}
\label{subsec:batch_normalization_description}
Ioffe and Szegedy~\cite{ioffe2015batch} proposed a batch normalization layer which has become an important component of modern day DNNs. Each batch normalization layer in the network calculates the mean and variance for each activation coming from the training data $X$, and normalizes each incoming sample $x$ as
\begin{equation}
 \hat{x} = \frac{x - {\expect{{X}}}}{\sqrt{{\var{{X}}} + \epsilon}} \cdot \gamma + \beta,
  \label{eq:bn}
\end{equation}
where $\gamma$ and $\beta$ are the scale and shift parameters, and $\epsilon$ is used for numerical stability. The expected value {\expect{{X}}} of the training statistics is estimated through the running mean,
\begin{equation}
  {\hat\mu_{k}} = (1-\rho)\cdot{\hat\mu_{k-1}} + \rho \cdot {\mu_{k}},
  \label{eq:running_mean}
\end{equation}
and the variance of the training statistics \var{{X}} is estimated through running variance,
\begin{equation}
  {\hat\sigma_{k}^2} = (1-\rho)\cdot{\hat\sigma_{k-1}^2} + \rho \cdot \sigma_{k}^2.
  \label{eq:running_var}
\end{equation}
Here, $\hat\mu$ and $\hat\sigma^2$ are the estimated mean and variance from the training data, whereas $\mu$ and $\sigma^2$ represent the mean and variance of the incoming batch. The hyperparameter $\rho$ is the momentum term (default $\rho=0.1$) and $k$ denotes each training step. Intuitively, $\rho$ can be thought of as the factor which controls how much the existing estimate of statistics is affected by the statistics of the incoming batch. A larger momentum value would essentially give more weight to the calculated statistics of the incoming batch. Empirically, it has been shown that batch normalization helps to train faster and also stabilize the training process\cite{santurkar2018does}. 

The behavior of batch normalization differs during training and testing as follows:

\paragraph{Training:}\hspace*{-3.5mm} During training, the batch normalization layer calculates the running mean and variance over the complete training set. The scale parameter $\gamma$ and shift parameter $\beta$ from Eq.~\eqref{eq:bn} are learned by back propagation. Running mean and variance is updated during each forward pass with the new batch statistics.

\paragraph{Testing:}\hspace*{-3.5mm} During inference, the running mean and variance of the batch normalization layer is fixed. Each new sample encountered during testing is normalized by using the population statistics calculated during training. 


\subsection{Dynamic Unsupervised Adaptation}
\label{subsec:method}
Let $\Phi_{\text{src}}$ be the network trained solely with source data $X_\text{src}$. Our goal is to adapt the trained model to out-of-distribution target data $X_\text{tar}$ in an unsupervised manner. The batch normalization layer performs consistently well when train and test data belong to a similar distribution~\cite{he2016deep,xie2017aggregated}. However, in many practical scenarios this is not the case. It has been shown that when out-of-distribution test data is encountered, batch normalization can hamper the performance significantly~\cite{li2016revisiting, nado2020evaluating, schneider2020improving, wang2020tent, wu2021rethinking, galloway2019batch}. One reason for the performance degradation is the misalignment of the activation distribution between training and out-of-distribution test data as shown in Figure~\ref{fig:sub1}. Thus, our adaptation process aligns the activation distribution between training and shifted test data as depicted in Figure~\ref{fig:sub2}.

\begin{figure*}
\centering
\begin{subfigure}{.5\textwidth}
  \centering
  \includegraphics[scale = 0.60]{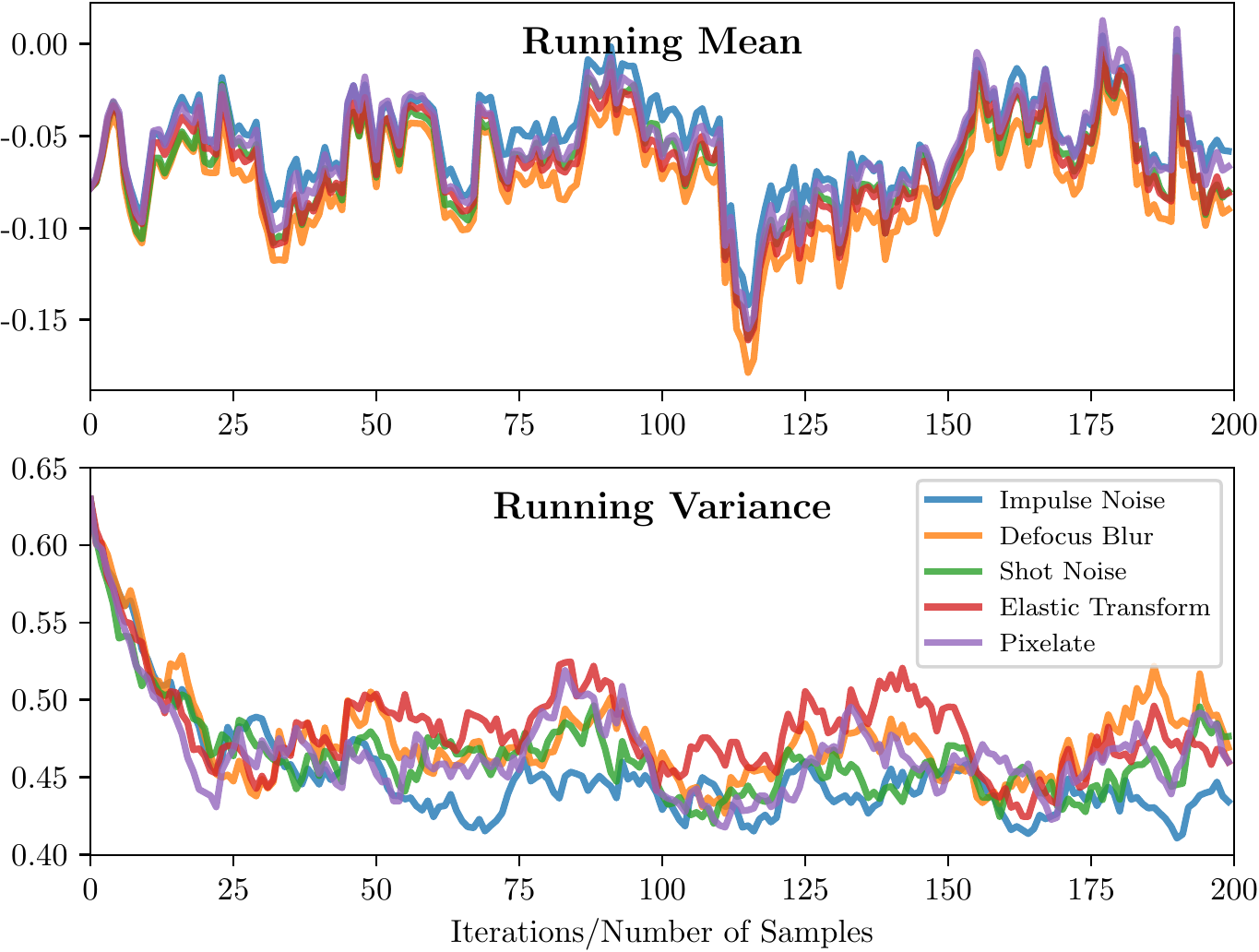}
  \caption{Default Parameters}
  \label{fig:running_mean_var_default}
\end{subfigure}%
\begin{subfigure}{.5\textwidth}
  \centering
  \includegraphics[scale = 0.60]{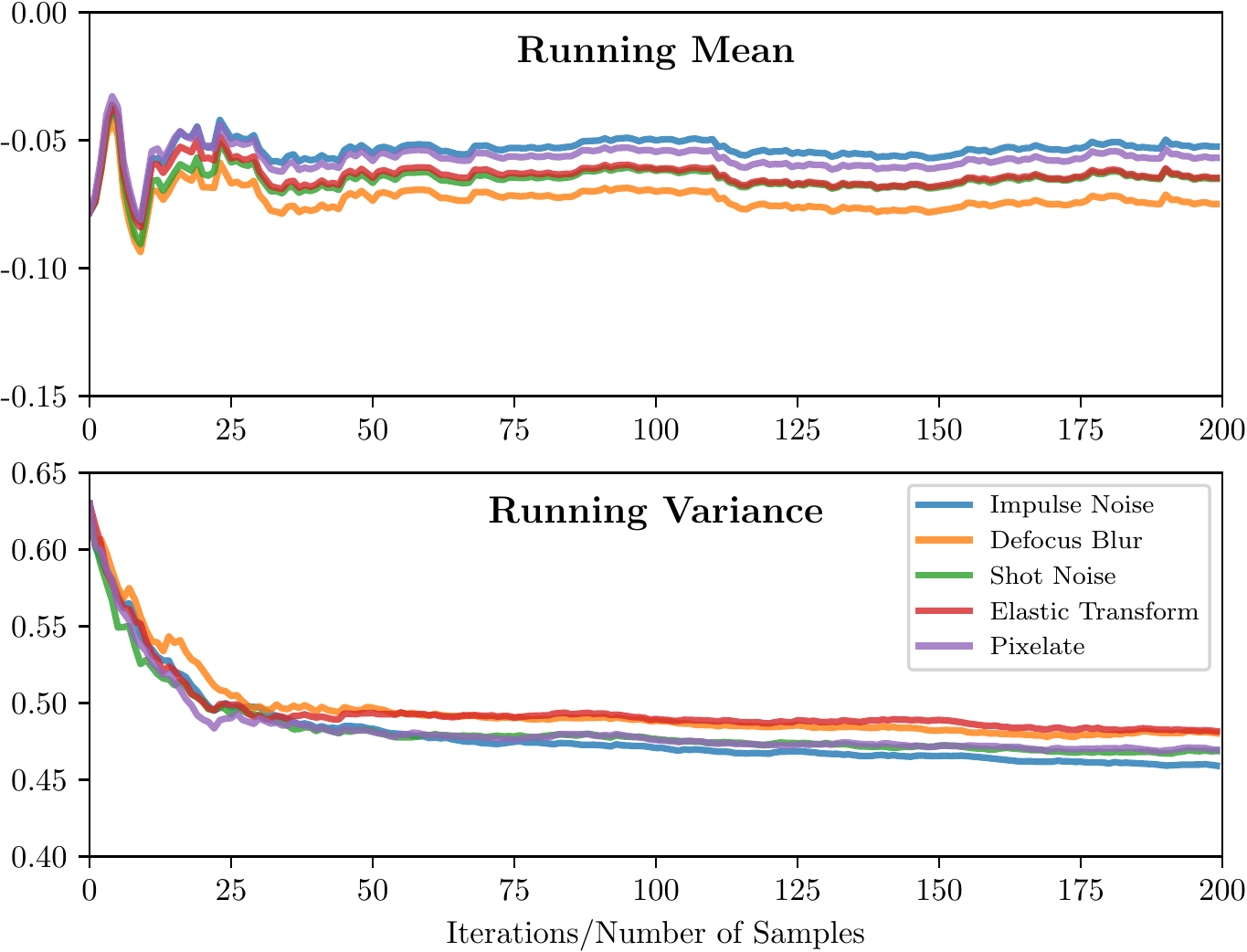}
  \caption{DUA}
  \label{fig:running_mean_var_DUA}
\end{subfigure}
\caption{Running mean and variance of a single channel from the last batch normalization layer for different corruptions in CIFAR-10C. a) Running mean and variance values at each adaptation iteration when default momentum parameters are used. The values are highly unstable which leads to unstable adaptation.
b) DUA proposes to use an adaptive momentum schema which leads to fast and stable convergence. This is because of the stability of the running mean and variance values. Initially, the distributions are far apart and thus, we want larger update steps (faster assimilation), whereas later on smaller update steps are beneficial.}
\label{fig:running_mean_var_comparison}
\end{figure*}

In our proposed adaptation schema, all the parameters of the network $\Phi_{\text{src}}$, other than the running mean and running variance are fixed. We only adapt the \expect{{X}} and \var{{X}} from Eq.~\eqref{eq:bn} to the new statistics of $X_\text{tar}$, by using the training statistics obtained from $X_\text{src}$ as a prior. The training statistics are updated by using one image after the other, \ie processing new examples of the (shifted) test data in a sequential manner as they arrive. The na\"ive approach would be to update the statistics by using Eqs.~\eqref{eq:running_mean} and~\eqref{eq:running_var} with a fixed momentum parameter $\rho$. However, as shown in Figure~\ref{fig:comparison_bn_mom}, such fixed momentum leads to a couple of problems: The adaptation performance is either unstable or converges rather slowly. This is because with the default parameters the adaptation of the running mean and variance is highly unstable, as shown in Figure~\ref{fig:running_mean_var_default}. Thus, for stable and fast convergence we adapt the momentum with each incoming sample. More formally, we update the mean and variance consecutively:
\begin{equation}
  {\hat\mu_{k}} = (1-(\rho_{k}+ \zeta))\cdot{\hat\mu_{k-1}} + (\rho_{k}+ \zeta) \cdot \mu_{k},
  \label{eq:adapt_mean_1}
\end{equation}
with 
\begin{equation}
   {\hat\mu_{0}} = {\hat\mu_{s}},\quad \rho_{k} = \rho_{k-1} \cdot \omega, \quad \rho_{0} = 0.1, 
  \label{eq:adapt_mean}
\end{equation}
and
\begin{equation}
  {\hat\sigma^2_{k}} = (1-(\rho_{k}+ \zeta))\cdot{\hat\sigma^2_{k-1}} + (\rho_{k}+ \zeta) \cdot \sigma^2_{k},
  \label{eq:adapt_var_1}
\end{equation}
with 
\begin{equation}
  {\hat\sigma^2_{0}} = {\hat\sigma^2_{s}}, \quad \rho_{k} = \rho_{k-1} \cdot \omega, \quad \rho_{0} = 0.1. 
  \label{eq:adapt_var}
\end{equation}

Here, $\omega$ $\in$ $(0,1)$, is the momentum decay parameter, whereas $\zeta$, with $0\,$\textless$\,\zeta\,$\textless$\,\rho_{0}$, is a constant and defines the lower bound of the momentum. As the momentum $\rho_k$ decays, the later samples will have a smaller impact. Our adaptive momentum scheme has a direct impact on how the running mean and variance are adapted. 
The adaptation becomes stable as compared to default parameters, which is visible in Figure~\ref{fig:running_mean_var_DUA}.   
   
Whenever we obtain a new sample from the (shifted) test distribution, we make a small batch by augmenting the incoming sample.
In particular, we use random horizontal flipping, random cropping and rotation. 
We take care that the augmentations we use do not correlate with the shifted test data in our experiments (\eg we do not augment with any of the corruptions in CIFAR-10/100C~\cite{hendrycks2019robustness}). An example of a batch we use for adaptation is provided in the supplemental material. Throughout our evaluations, we found that making a small batch from a single image stabilizes the adaptation process and improves the results, although this is not strictly necessary for our adaptation scheme to work. The effect of batches and augmentations is analyzed in our ablation study in Sec.~\ref{sec:ablation_studies}.

\begin{table*}
  \centering
  \small
  \begin{tabular}{c|ccccccccccccccc|c}
  \toprule
    & gaus & shot & impul & defcs & gls & mtn & zm & snw & frst & fg & brt & cnt & els & px & jpg & mean\\
    \midrule
    
      Source &            {67.7} &            {63.1} &            {69.9} &            {55.3} &            {56.6} &            {42.2} &            {50.1} &            {31.6} &            {46.3} &            {39.1} &            {17.1} &            {74.6} &            {34.2} &            {57.9} &            {31.7} &  49.2 \\
    TTT &            {45.6} &            {41.8} &            {50.0} &            {21.8} &            {46.1} & {\bfseries{23.0}} &            {23.9} &            {29.9} &            {30.0} &            {25.1} & {\bfseries{12.2}} & {\bfseries{23.9}} & {\bfseries{22.6}} &            {47.2} & {\bfseries{27.2}} &  31.4 \\
   NORM &            {44.6} &            {43.7} &            {49.1} &            {29.4} &            {45.2} &            {26.2} &            {26.9} &            {25.8} &            {27.9} &            {23.8} &            {18.3} &            {34.3} &            {29.3} &            {37.0} &            {32.5} &  32.9 \\
    DUA & {\bfseries{34.9}} & {\bfseries{32.6}} & {\bfseries{42.2}} & {\bfseries{18.7}} & {\bfseries{40.2}} &            {24.0} & {\bfseries{18.4}} & {\bfseries{23.9}} & {\bfseries{24.0}} & {\bfseries{20.9}} &            {12.3} &            {27.1} &            {27.2} & {\bfseries{26.2}} &            {28.7} &  \textbf{26.8} \\
    
    \midrule
    Source &            {28.8} &            {22.9} &            {26.2} &            \phantom{0}{9.5} &            {20.6} &           {10.6} &            \phantom{0}{9.3} &            {14.2} &            {15.3} &            {17.5} &            \phantom{0}{7.6} &            {20.9} &            {14.7} &            {41.3} &            {14.7} & 18.3 \\
  TENT &            {15.8} &            {13.5} &            {18.7} &            \phantom{0}{8.1} &            {18.7} & {\phantom{0}\bfseries{9.1}} &            \phantom{0}{8.0} & {\bfseries{10.3}} & {\bfseries{10.8}} & {\bfseries{11.7}} &            \phantom{0}{6.7} &            {11.6} &            {14.1} & {\bfseries{11.7}} &            {15.2} & 12.3 \\
   DUA & {\bfseries{15.4}} & {\bfseries{13.4}} & {\bfseries{17.3}} & {\phantom{0}\bfseries{8.0}} & {\bfseries{18.0}} & {\phantom{0}\bfseries{9.1}} & {\phantom{0}\bfseries{7.7}} &            {10.8} & {\bfseries{10.8}} &            {12.1} & {\phantom{0}\bfseries{6.6}} & {\bfseries{10.9}} & {\bfseries{13.6}} &            {13.0} & {\bfseries{14.3}} & \textbf{12.1} \\
\bottomrule
  \end{tabular}
  \caption{Top-1 Classification Error (\%) for each corruption in CIFAR-10C at the highest severity (Level 5). \emph{Source} shows the results from the same model trained on the clean train set and tested on the corrupted test set. For a fair comparison with TTT and NORM, we use ResNet-26 (top), while for TENT, we use the WRN-40-2 (bottom) from their official implementation. Smallest error is shown in bold.}
  \label{tab:cifar_10_results_level-5}
\end{table*}
\begin{table*}
  \centering
  \small
  \begin{tabular}{c|ccccccccccccccc|c}
  \toprule
    & gaus & shot & impul & defcs & gls & mtn & zm & snw & frst & fg & brt & cnt & els & px & jpg & mean\\
    \midrule
  
    Source &            {89.5} &            {88.8} &            {95.5} &            {68.4} &            {83.3} &            {65.0} &            {63.5} &            {62.4} &            {74.9} &            {70.3} &            {42.9} &            {83.0} &            {61.1} &            {84.4} &            {65.5} & 73.2 \\
   TTT &            {83.8} &            {83.0} &            {86.8} &            {59.9} &            {77.7} &            {57.9} &            {59.2} &            {61.5} &            {70.6} &            {70.5} &            {44.5} &            {69.8} &            {56.5} &            {80.2} & {\bfseries{60.3}} & 68.1 \\
  NORM &            {72.5} &            {72.7} &            {77.1} &            {48.6} &            {69.3} & {\bfseries{49.7}} &            {47.9} &            {59.5} &            {59.7} &            {58.4} &            {41.8} & {\bfseries{53.1}} &            {58.8} &            {57.3} &            {67.7} & 59.6 \\
   DUA & {\bfseries{67.9}} & {\bfseries{67.3}} & {\bfseries{72.6}} & {\bfseries{47.9}} & {\bfseries{66.1}} &            {51.6} & {\bfseries{46.6}} & {\bfseries{58.1}} & {\bfseries{57.6}} & {\bfseries{54.4}} & {\bfseries{41.3}} &            {58.6} & {\bfseries{55.3}} & {\bfseries{53.3}} &            {60.7} & \textbf{57.3} \\
    
    \midrule
    Source&65.7&60.1&59.1&32.0&51.0&33.6&32.4&41.4&45.2&51.4&31.6&55.5&40.3&59.7&42.4&46.7
\\
TENT&\textbf{40.3}&\textbf{39.9}&41.8&\textbf{29.8}&\textbf{42.3}&\textbf{31.0}&30.0&\textbf{34.5}&\textbf{35.2}&\textbf{39.5}&\textbf{28.0}&\textbf{33.9}&\textbf{38.4}&\textbf{33.4}&41.4&\textbf{36.0}
\\
DUA&42.2&40.9&\textbf{41.0}&30.5&44.8&32.2&\textbf{29.9}&38.9&37.2&43.6&29.5&39.2&39.0&35.3&\textbf{41.2}&37.6
\\
    \bottomrule
  \end{tabular}
  \caption{Top-1 Classification Error (\%) for each corruption in CIFAR-100C at the highest severity (Level 5).}
  \label{tab:cifar_100_results_level-5}
\end{table*}

\begin{table*}
  \centering
  \small
  \begin{tabular}{c|ccccccccccccccc|c}
    \toprule
    
    & gaus & shot & impul & defcs & gls & mtn & zm & snw & frst & fg & brt & cnt & els & px & jpg & mean\\
    \midrule
Source &            {98.4} &            {97.7} &            {98.4} &            {90.6} &            {93.4} &            {89.8} &            {81.8} &            {89.5} &            {85.0} &            {86.3} &            {51.1} &            {97.2} &            {85.3} &            {76.9} &            {71.7} & 86.2 \\
   TTT &            {96.9} &            {95.5} &            {96.5} &            {89.9} &            {93.2} &            {86.5} &            {81.5} &            {82.9} &            {82.1} &            {80.0} &            {53.0} & {\bfseries{85.6}} &            {79.1} &            {77.2} &            {74.7} & 83.6 \\
  NORM & {\bfseries{87.1}} &            {89.6} &            {90.5} & {\bfseries{87.6}} &            {89.4} & {\bfseries{80.0}} & {\bfseries{71.9}} & {\bfseries{70.6}} &            {81.5} & {\bfseries{66.9}} &            {47.8} &            {89.8} &            {73.5} & {\bfseries{64.2}} &            {68.5} & \textbf{77.3} \\
   DUA &            {89.4} & {\bfseries{87.6}} & {\bfseries{88.1}} &            {88.0} & {\bfseries{88.6}} &            {84.7} &            {74.3} &            {77.8} & {\bfseries{78.4}} &            {68.6} & {\bfseries{45.6}} &            {95.9} & {\bfseries{72.2}} &            {66.5} & {\bfseries{67.4}} & 78.2 \\
    \bottomrule
  \end{tabular}
  \caption{Top-1 Classification Error (\%) for each corruption in ImageNet-C at the highest severity (Level 5). \emph{Source} refers to results obtained from a model pre-trained on the original ImageNet and tested on the corrupted test sets. All results are obtained using a ResNet-18 backbone. Smallest error is shown in bold.}
  \label{tab:ImageNet_results}
\end{table*}

\section{Results}
\label{sec:results}
In the following, we evaluate DUA on a variety of tasks and benchmarks. First, we summarize the datasets. Next, we introduce the approaches to which we compare. Lastly, we present our detailed results. 

\subsection{Benchmarks and Tasks}

\paragraph{CIFAR-10/100C:}CIFAR-10C and CIFAR-100C~\cite{hendrycks2019robustness} are image classification benchmarks to test a model's robustness w.r.t. covariate shifts. These benchmarks add different corruptions to the original test set of CIFAR-10/100~\cite{krizhevsky2009learning} at 5 severity levels. Following the common protocol~\cite{wang2020tent, nado2020evaluating, sun2020test}, we evaluate on 15 types of corruptions.

\paragraph{ImageNet-C:} Similar to the CIFAR-10/100C benchmarks, ImageNet-C~\cite{hendrycks2019robustness} is also an image classification dataset introducing different corruptions at several severity levels to the original test set of ImageNet~\cite{deng2009imagenet}.




\paragraph{KITTI:} To test DUA's adaptation capabilities on the task of object detection for autonomous vehicles we use the well-known KITTI~\cite{geiger2013vision} dataset. Further, we also use the KITTI-Rain and KITTI-Fog datasets~\cite{halder2019physics} to test the adaptation performance of a KITTI pre-trained model in degrading weather.



\subsection{Baselines}
We compare our DUA against the following approaches: 

\begin{itemize}
    \item \textbf{Source}: denotes the results of the corresponding baseline model trained only on source data, \ie without any adaptation to the test data. 
    \item \textbf{TTT}: Test Time Training (TTT)~\cite{sun2020test} adapts the network parameters by using an auxiliary task on each (out-of-distribution) data sample before testing it. 
    \item \textbf{NORM}~\cite{schneider2020improving ,nado2020evaluating}: ignores the train statistics completely and recalculates the batch normalization statistics on the entire test set, leveraging larger batch sizes. 
    \item \textbf{TENT}: Test time entropy minimization (TENT)~\cite{wang2020tent} recalculates the batch normalization statistics and additionally modifies the scale and shift parameters ($\gamma$ and $\beta$) of the batch normalization layers through back propagation.
    They obtain the gradients by calculating the prediction entropy on large batches from the out-of-distribution test data.
\end{itemize}
\subsection{Experiments}
In this section, we provide a description of all the results obtained on different datasets and benchmarks. We test our DUA during slight distribution shifts and severe domain shifts as well. For our results we always use less than 1\% of unlabeled test data and adapt on each incoming sample in a sequential manner (the exact numbers of test samples used for adaptation in our experiments are listed in the supplemental material). Results for all other baselines have been obtained by adapting on the complete test set as stated in their original papers. Note, we also do not need to control for shuffling of the test set in contrast to other approaches~\cite{wang2020tent, schneider2020improving, nado2020evaluating}. For adaptation, unless stated otherwise, we fix the momentum decay parameter $\omega=0.94$, and the lower bound $\zeta=0.005$. 
Further details for all the experiments are provided in the supplemental material. 
For reproducibility, code for DUA is available at this repository: \url{https://github.com/jmiemirza/DUA}
\subsection*{CIFAR-10/100C}
\begin{table*}
    \centering
    \small
    \subfloat[KITTI $\rightarrow$ KITTI-Fog\label{tab:kitti_fog_rain:fog}]{
\begin{tabular}{c|ccc}
    \toprule
  & Car & Pedestrian & Cyclist \\
    \midrule
    
    Source only & 30.9 &34.1 &16.2\\
    DUA & 51.4 &48.5 &33.1\\
    Fully Supervised & 71.3 &64.5 &63.2\\
    \bottomrule
  \end{tabular}}\hspace{1cm}
  \subfloat[KITTI $\rightarrow$ KITTI-Rain\label{tab:kitti_fog_rain:rain}]{
\begin{tabular}{c|ccc}
    \toprule
     & Car & Pedestrian & Cyclist \\
    \midrule
    
    Source only & 80.7 &66.7 &54.6\\
    DUA & 86.3 &70.3 &66.7\\
    Fully Supervised & 92.3 &76.1 &78.2\\
    \bottomrule
  \end{tabular}}
\caption{Results for KITTI pre-trained YOLOv3 tested on rain and fog datasets. We report the Mean Average Precision (mAP@50). a)~Results for the most severe fog level,  \ie $30$m visibility. b) Results for the most severe rain level, \ie $200$mm/hr rain intensity.}
\label{tab:kitti_fog_rain}
\end{table*}
For a fair comparison to TTT and NORM we use ResNet-26~\cite{he2016deep} and follow the parametrization in their official implementations\footnote{TTT: \href{https://github.com/yueatsprograms/ttt_cifar_release}{https://github.com/yueatsprograms/ttt\_cifar\_release}}$^{\text{,}}$\footnote{NORM: \href{https://github.com/bethgelab/robustness}{https://github.com/bethgelab/robustness}}. Similarly, for TENT we use the Wide-ResNet-40-2~\cite{zagoruyko2016wide} from their official implementation\footnote{TENT: \href{https://github.com/DequanWang/tent}{https://github.com/DequanWang/tent}}.

Table~\ref{tab:cifar_10_results_level-5} shows the results for the highest severity level on CIFAR-10C. Note that we achieve a new state-of-the-art. All other approaches use the complete test set and most also use larger batch sizes and control shuffling of test data.  Results on CIFAR-100C are listed in Table~\ref{tab:cifar_100_results_level-5}.
Here, DUA outperforms TTT and NORM while being competitive with TENT.
Results for lower severity levels are provided in the supplemental material, demonstrating that DUA provides strong results for less severe corruptions as well. 

\subsection*{ImageNet-C}
For evaluations on ImageNet-C we take an off-the-shelf pre-trained ResNet-18 from PyTorch~\cite{NEURIPS2019_9015}. Table~\ref{tab:ImageNet_results} shows the top-1 error for the highest severity level. DUA performs on-par with all the baselines for ImageNet-C.
Results for lower severity levels are provided in the supplemental.

\subsection*{Object Detection}
We also test our approach for object detection and show considerable improvement. We conduct these experiments with YOLOv3~\cite{redmon2018yolov3}. However, our approach could also be applied to other base architectures such as~\cite{ren2015faster, lin2017focal, liu2016ssd, tan2020efficientdet}, which use batch normalization.

For evaluating our approach on object detection we consider the two following scenarios: 

\begin{itemize}
    \item Evaluation during covariate shifts; These evaluations are performed to adapt to rain and fog conditions. 
    \item Evaluation during domain shifts; These evaluations test for domain adaptation between datasets. 
\end{itemize}

In degrading weather, a sharp drop in performance of present day object detectors has been noted~\cite{mirza2021robustness,michaelis2019benchmarking}. Our goal is to dynamically adapt a detector trained on clear weather data to degrading weather conditions.
Adaptation results for the most severe forms of fog and rain augmented on KITTI are shown in Table~\ref{tab:kitti_fog_rain:fog} and~\ref{tab:kitti_fog_rain:rain}, respectively.
For fog, the mean improvement across all commonly evaluated classes (\ie car, pedestrian and cyclist) over the source model is 17.7\% mAP. Similarly, we also achieve notable improvements while adapting to rain. Here, the mean improvement over the source model is 7.1\% mAP. Additional results for varying severity of fog and rain are provided in the supplemental material.

\subsection*{Additional Results}


We also demonstrate the benefits of DUA on several other datasets and adaptation tasks in the supplemental material.
In particular, we evaluate DUA on:

\paragraph{Digit Recognition:} DUA can successfully be used for domain adaptation across datasets which we demonstrate for the task of digit recognition.
In particular, we use MNIST~\cite{lecun1998gradient} and USPS~\cite{hull1994database}, which are datasets consisting of handwritten digits. Additionally, we use SVHN~\cite{netzer2011reading}, a dataset containing house numbers obtained from Google street view images. 

\paragraph{Office-31}\hspace{-2mm}{\cite{saenko2010adapting}\textbf{:}}\hspace{3mm} is a visual domain adaptation dataset for object classification, containing 31 categories of common objects found in an office environment, captured in three different settings. These include images captured by Webcam, DSLR and gathered from Amazon. We test for domain adaptation across all three settings.

\paragraph{VIS-DA:} The Visual Domain Adaptation~\cite{peng2017visda} dataset (VIS-DA) is a large scale image recognition dataset which contains 12 classes. The training set consists of synthetically rendered images. The test set consists of real images cropped from the MS-COCO dataset~\cite{lin2014microsoft}.

\paragraph{SODA10M:} The large scale object detection dataset for autonomous vehicles SODA10M~\cite{han2021soda10m} provides data captured during day and night. We test for adaptation from day to night. Further, we test for domain adaptation between KITTI and SODA10M.  

\section{Ablation Studies}
\label{sec:ablation_studies}
In this section we present detailed ablation studies in order to examine our approach more closely.
\subsection{Sample Order Does Not Matter}
To understand if the ordering of the incoming samples for adaptation holds any significance, we run DUA for 300 independent runs (on CIFAR-10C) and randomly shuffle the test set in each run. The initial, source-only, mean error is $49.2\,\%$. The largest standard deviation over the 300 runs occurs after adaptation on 5 samples, where we achieve $36.4\pm0.4$. Already after 25 samples, we achieve $28.3\pm0.19$. The performance saturates after 100 samples and we achieve $27.2\pm0.09$ (a detailed plot is provided in supplemental material).
Thus, the performance of DUA is stable across all independent runs with very little deviation. These results are important to understand that DUA can be used with any arrangement of incoming data.   

\subsection{Continuous Dynamic Adaptation}
\begin{figure}
    \centering
    \includegraphics[scale=0.5]{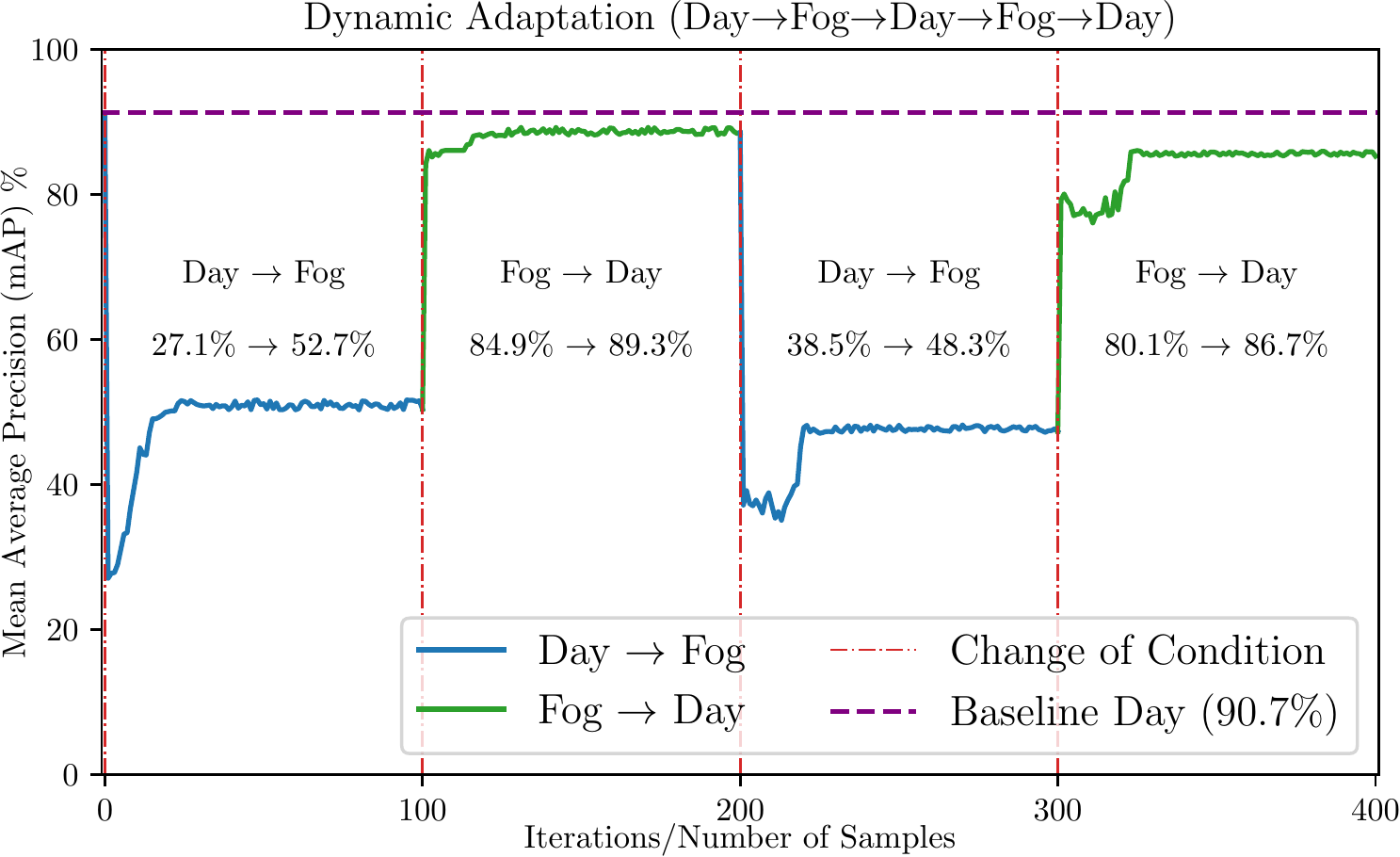}
    \caption{Dynamic adaptation scenario for DUA. We let a KITTI-pretrained model adapt to fog and then back to  the original KITTI dataset for two cycles in order to show how well DUA can dynamically adapt to changing weather conditions.
    }
    \label{fig:continuous_dynamic_adaptation}
\end{figure}

In order to understand how good can DUA handle the real-life scenarios where different weather conditions can occur interchangeably, we test DUA for such a scenario on KITTI-Fog: day$\to$fog$\to$day$\to$\emph{etc.} Results are shown in Fig.~\ref{fig:continuous_dynamic_adaptation}.
Note the only minor drop on the source domain (\eg 5.8\% worse than the \emph{day baseline} at iteration 100) and that we can quickly recover the same performance again. This shows that despite not being an incremental learning approach, DUA still remembers information from the previous domain. We conjecture that this is because learned weights are not changed during adaptation. DUA only shifts the activation distributions at test time.

\subsection{Ablating Batch Normalization Layers}
\begin{figure}
    \centering
    \includegraphics[scale=0.5]{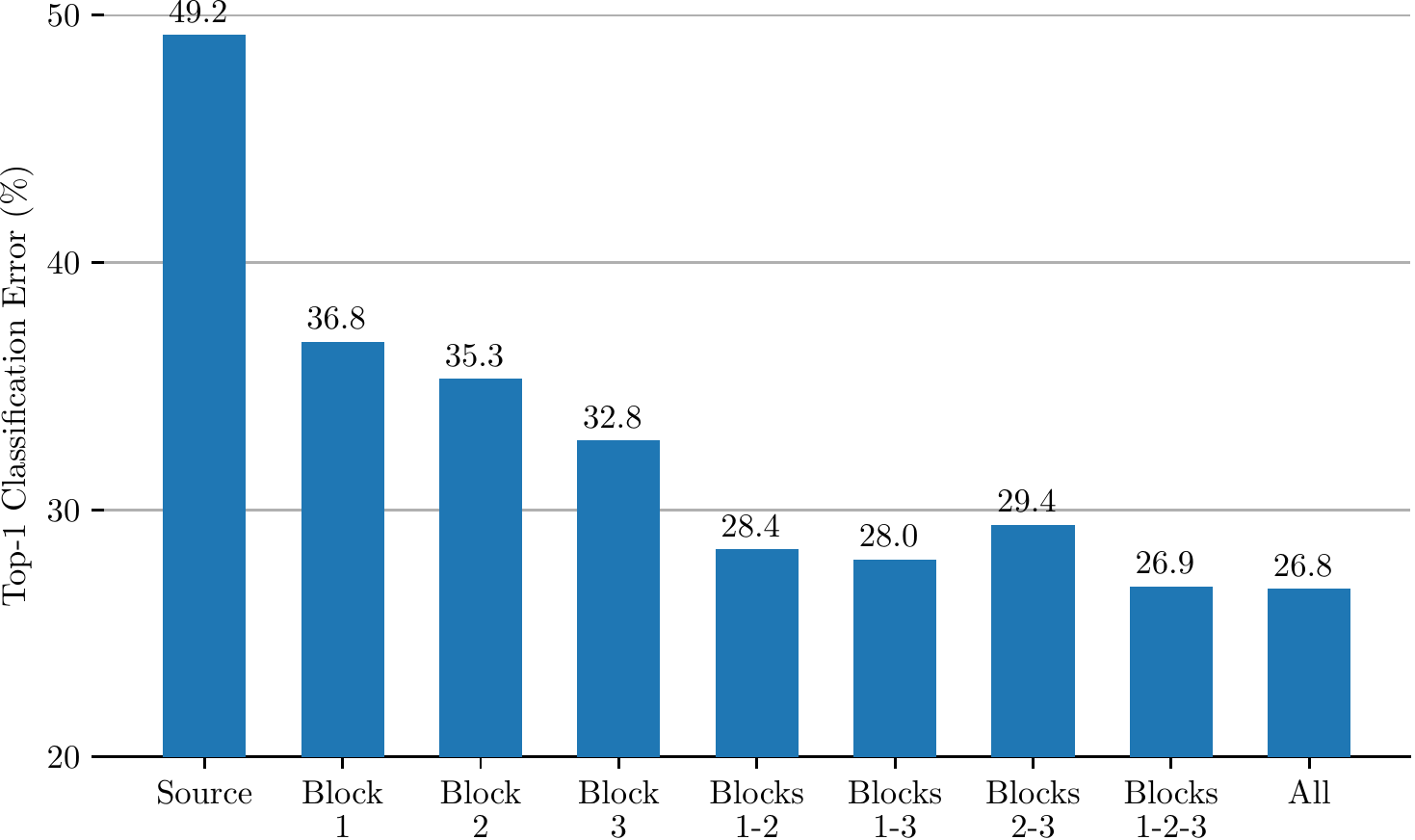}
    \caption{Results on CIFAR-10C after adapting the batch normalization layers of specific ResNet-26 blocks. `All' refers to adapting all batch normalization layers. This includes the last batch normalization layer after the three ResNet blocks.}
    \label{fig:adapting_diff_bn}
\end{figure}

We investigate the effect of adapting only selected batch normalization layers in Figure~\ref{fig:adapting_diff_bn}. For this, we adapt batch normalization layers of specific~ResNet-26 blocks while keeping all others fixed. As can be seen from the plot, the best performance is obtained by adapting all batch normalization layers in the architecture. Individual improvements are slightly larger at later batch normalization layers. 

\subsection{Effect of Augmentation}
\label{subsec:effect_of_aug}
\begin{figure}
    \centering
    \includegraphics[scale=0.5]{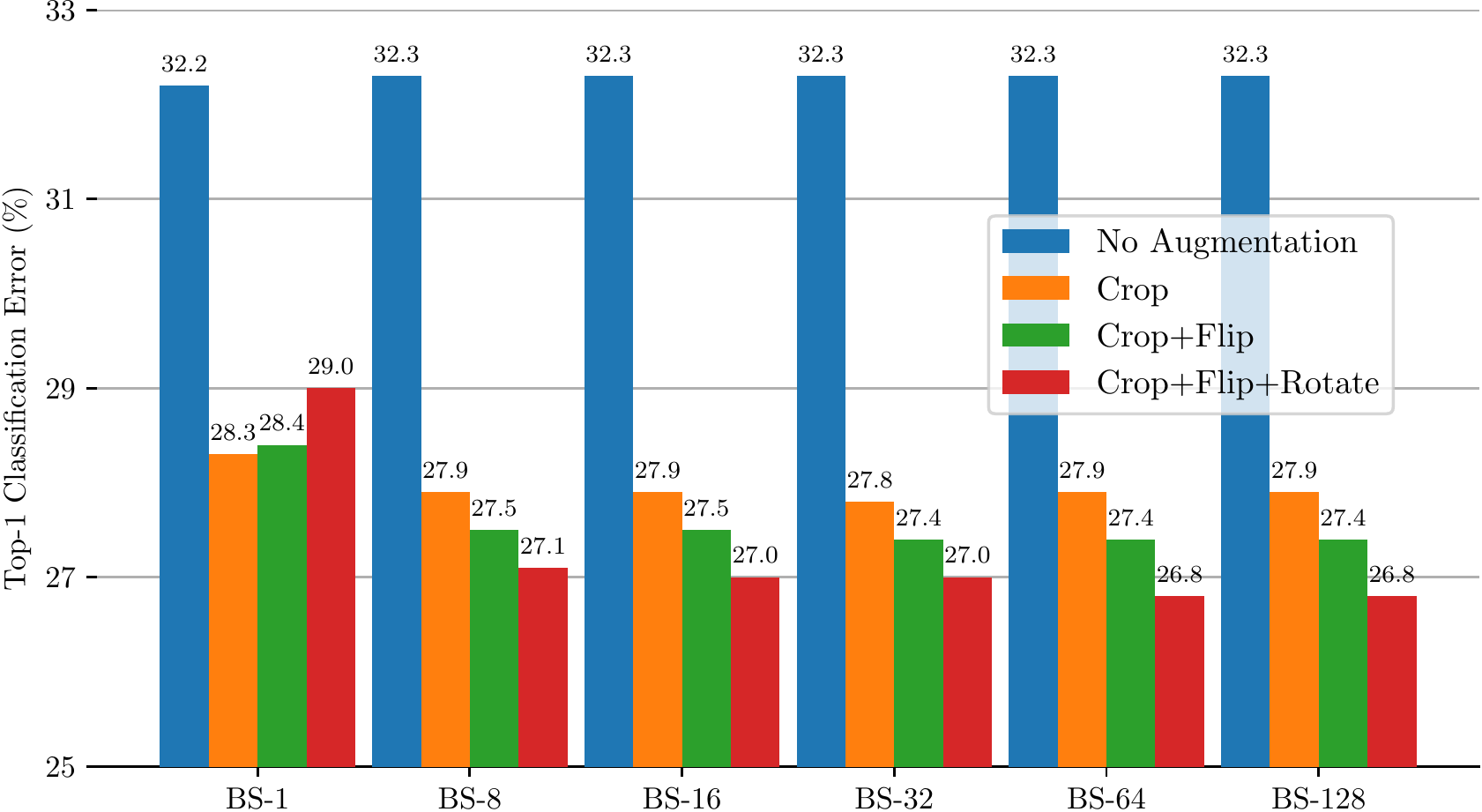}
    \caption{Effects of different batch sizes and augmentations on CIFAR-10C with ResNet-26.
    DUA makes a batch out of each sequential image by using different random augmentations. 
    Note that the source-only error (without adaptation) is \textbf{49.2\%}. 
    }
    \label{fig:aug_bs_ablation}
\end{figure}

As explained in Section~\ref{subsec:method}, we form a small batch of augmented versions from each incoming sample. In Figure~\ref{fig:aug_bs_ablation}, we ablate different augmentations and batch sizes to study their effects. Apart from providing stability to our adaptation procedure, making a small batch and augmenting it randomly also provides further improvements. For our experiments, we form a batch of size 64 from each incoming image by augmenting it. However, even a batch size of only 8 suffices to benefit from DUA (with only minor sacrifice in performance). 
\section{Conclusion}
\label{sec:discussion_and_conclusion}



We have shown that even slight distribution shifts between train and test data can greatly hamper the performance of present day neural networks.
We address this limitation by our DUA, which adapts the statistics of a trained model in a sequential manner on each unlabeled sample coming from out-of-distribution test data.
To ensure fast and stable adaptation, we introduce an adaptive momentum scheme. DUA does not require access to training data but only needs a fraction of test data to achieve competitive results to strong baselines. Extensive experimentation on a variety of challenging benchmarks and tasks demonstrate the utility of our method on a broad range of batch normalization-based architectures. 
Since we can dynamically adapt to shifting distributions at a minimal computational overhead, DUA is also well-suited for both real-time systems and embedded devices. 

\paragraph{Acknowledgments}
\label{sec:acknowledgement}
We gratefully acknowledge the financial support by the Austrian Federal Ministry for Digital and Economic Affairs, the National Foundation for Research, Technology and Development and the Christian Doppler Research Association. This work was also partially funded by the Austrian Research Promotion Agency~(FFG) under the project High-Scene~(884306).


{\small
\bibliographystyle{ieee_fullname}
\bibliography{dua}
}
\begin{appendices}
In the following, we summarize our evaluation details to support reproducibility (Section~\ref{sec:eval-details}) and provide additional detailed results (Section~\ref{sec:results-plus-plus}).
\section{Evaluation Details}
\label{sec:eval-details}
\paragraph*{Number of Samples:}
DUA requires only a tiny fraction of the (unlabeled) test set to achieve competitively strong results.
The exact numbers are listed in Table~\ref{tab:num_samples}. Note that after adaptation on these specific number of samples, the adaptation performance always saturates. 
We conducted all experiments on a single NVIDIA\textsuperscript{\textregistered} GeForce\textsuperscript{\textregistered} RTX 3090.
\begin{table*}
\small
  \centering
  \begin{tabular}{c|ccc}
    \toprule
    \multirow{2}{*}{Dataset} & Total no. of samples in & No. of samples used & \% of samples used \\
            &test set& for adaptation &  for adaptation \\
    \midrule
    
    CIFAR-10C~\cite{hendrycks2019robustness} & 10000 &\phantom{0}80 &0.8\%\\
    CIFAR-100C~\cite{hendrycks2019robustness} & 10000 &\phantom{0}80 &0.8\%\\
    ImageNet-C~\cite{hendrycks2019robustness} & 50000 &100 &0.2\%\\
    MNIST~\cite{lecun1998gradient} & 10000 &\phantom{0}30 &0.3\%\\
    SVHN~\cite{netzer2011reading} & 26032 &\phantom{0}30 &0.1\%\\
    USPS~\cite{hull1994database} & \phantom{0}2007 &\phantom{0}20 &0.9\%\\
    Office-31~\cite{saenko2010adapting} & \phantom{0}4110 &\phantom{0}40 &0.9\%\\
    VIS-DA~\cite{peng2017visda} & \phantom{0}5534 &\phantom{0}40 &0.7\%\\
    KITTI~\cite{geiger2013vision} & \phantom{0}3741 &\phantom{0}25 &0.6\%\\
    KITTI-Rain~\cite{halder2019physics} & \phantom{0}3741 &\phantom{0}25 &0.6\%\\
    KITTI-Fog~\cite{halder2019physics} & \phantom{0}3741 &\phantom{0}25 &0.6\%\\
    SODA-10M~\cite{han2021soda10m} & \phantom{0}4991 &\phantom{0}30 &0.6\%\\
    SODA10M-Day~\cite{han2021soda10m} & \phantom{0}3347 &\phantom{0}30 &0.8\%\\
    SODA10M-Night~\cite{han2021soda10m} & \phantom{0}1644 &\phantom{0}15 &0.9\%\\

    \bottomrule
  \end{tabular}
  \caption{Available number of samples in the test sets \emph{vs.} number of samples used for adaptation. Note that all datasets except for KITTI have dedicated test sets. On KITTI, we divide the training set of 7481 images into 3740 train and 3741 test images. }
  \label{tab:num_samples}
\end{table*}

\begin{figure*}
    \centering
    \includegraphics[scale=0.15]{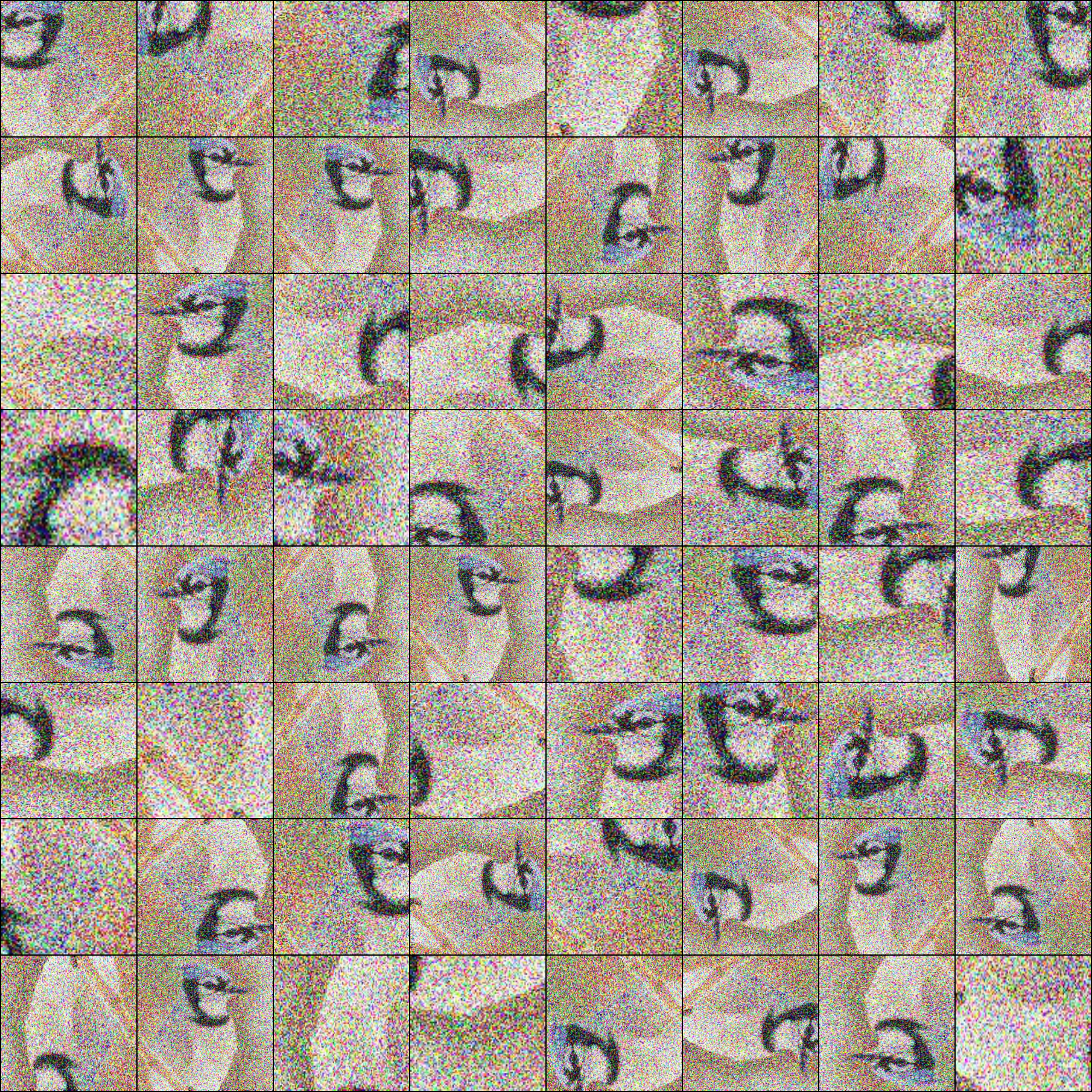}
    \caption{Example of a batch which we create from a single image by augmenting it randomly. The input sample is taken from ImageNet-C~\cite{hendrycks2019robustness} (Level 5) Gaussian Noise Corruption.}
    \label{fig:example_of_batch}
\end{figure*}

\paragraph*{Batch Augmentations:}
As evaluated in our ablation study (main manuscript Section 5), augmentations help to further improve the adaptation performance. In our experiments we use random cropping, random horizontal flipping and rotating by specific angles ($0, 90, 180, 270$ degrees). Figure~\ref{fig:example_of_batch} shows an exemplary batch of size 64.  

\paragraph*{Corruption Benchmarks:}
To train the ResNet-26~\cite{he2016deep} backbone on CIFAR-10/100~\cite{krizhevsky2009learning}, we use a batch size of 128. We train the model for 150 epochs and use Stochastic Gradient Descent (SGD) as optimizer with learning rate 0.1, momentum 0.9 and weight decay $5\cdot10^{-4}$. We use the multi-step learning rate scheduler from PyTorch with milestones at $75$ and $125$ and set its $\gamma=0.1$. For training we use random cropping and horizontal flipping as augmentations.

\paragraph*{Domain Adaptation for Classification:}
We use an ImageNet-pretrained ResNet-18 from PyTorch and finetune it on the Office-31~\cite{saenko2010adapting} train split.
For this finetuning, we train the model for 100 epochs with SGD, initial learning rate $0.1$, momentum $0.9$ and weight decay $5\cdot10^{-4}$. We use the PyTorch multi-step learning rate scheduler with milestones at $60$ and $75$ and set its $\gamma=0.1$.
For VIS-DA~\cite{peng2017visda} we use a ResNet-50, while all other settings are kept the same as for Office-31.
We again use random cropping and horizontal flipping as augmentations during training.

\paragraph*{Digit Recognition:}
For these experiments we use a simple architecture, consisting of 2 convolution layers, 2 linear layers and 2 batch normalization layers. We also use 2 dropout layers in the classification head for regularization with dropout probabilities $p=0.25$ and $p=0.5$, respectively. For each of the datasets, we train this model using ADADELTA~\cite{zeiler2012adadelta} for 15 epochs and use a batch size of 64.

\paragraph*{Object Detection:}
For all our object detection experiments we use a YOLOv3~\cite{redmon2018yolov3} pretrained on MS-COCO~\cite{lin2014microsoft}. Then, we retrain the model for 100 epochs on each of the training splits of the respective datasets. We use a batchsize of 20, while all other optimization settings and augmentation routines for training are taken from the PyTorch implementation\footnote{\href{https://github.com/ultralytics/yolov3}{https://github.com/ultralytics/yolov3}} of YOLOv3.

\section{Additional Ablation Results}
\label{sec:results-plus-plus}
\subsection{Sample Order Does Not Matter}
\begin{figure}
    \centering
    \includegraphics[width=8cm, height=6cm]{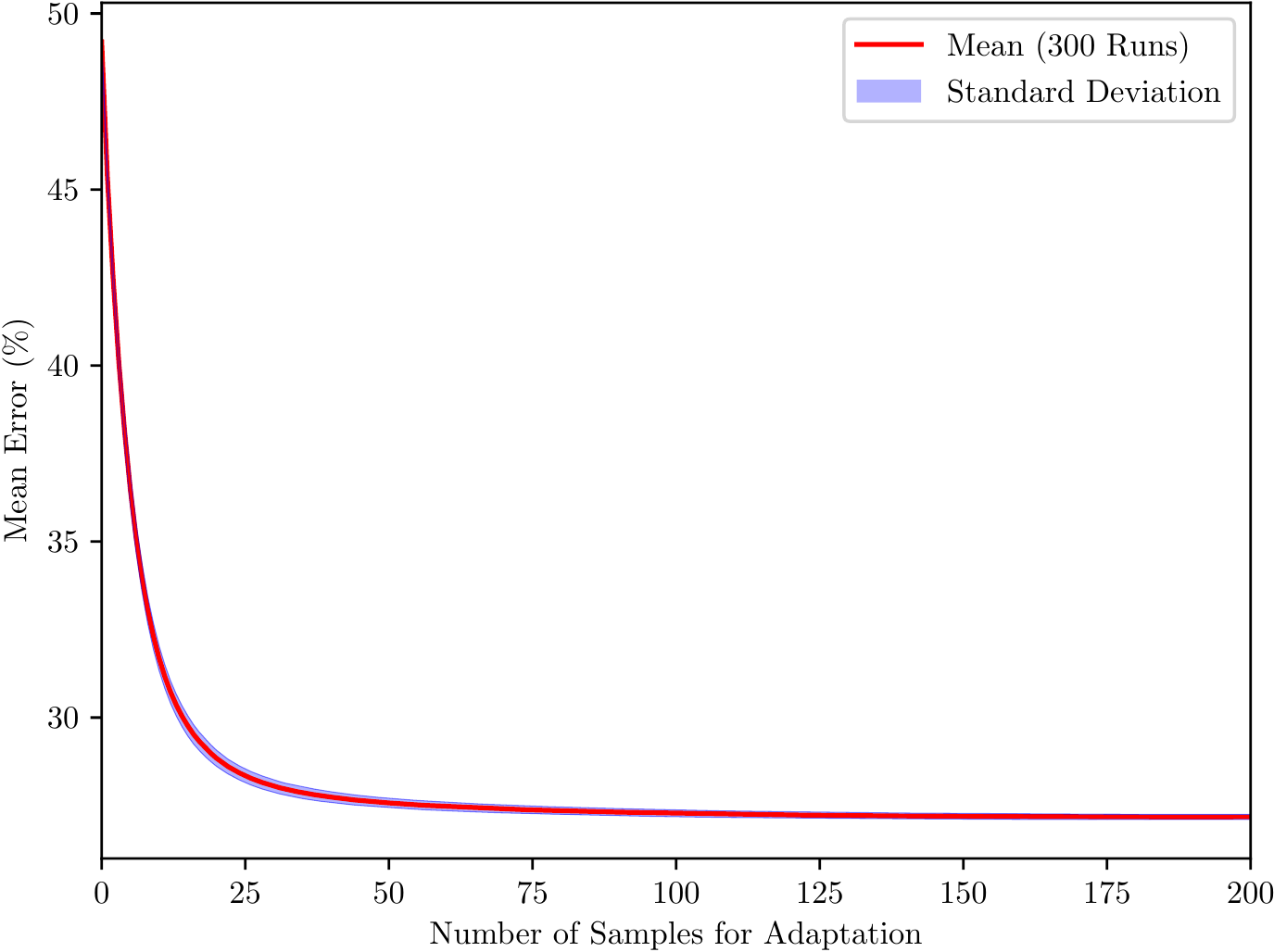}
    \caption{Adaptation results on CIFAR-10C for 300 runs with $\omega = 0.94$, $\rho_{0}=0.1$. For each run we randomly shuffle the corrupted test sets. We plot the mean error (on 15 corruptions) and the standard deviation after adaptation on each incoming sample.}
    \label{fig:random_runs}
\end{figure}

As described in the main manuscript (Section 5), the sample order does not matter. Here, we include the detailed plot for all the 300 runs in Figure~\ref{fig:random_runs}, which shows that DUA is consistently stable across all runs.   

\begin{figure}
    \centering
    \includegraphics[width=8cm, height=6cm]{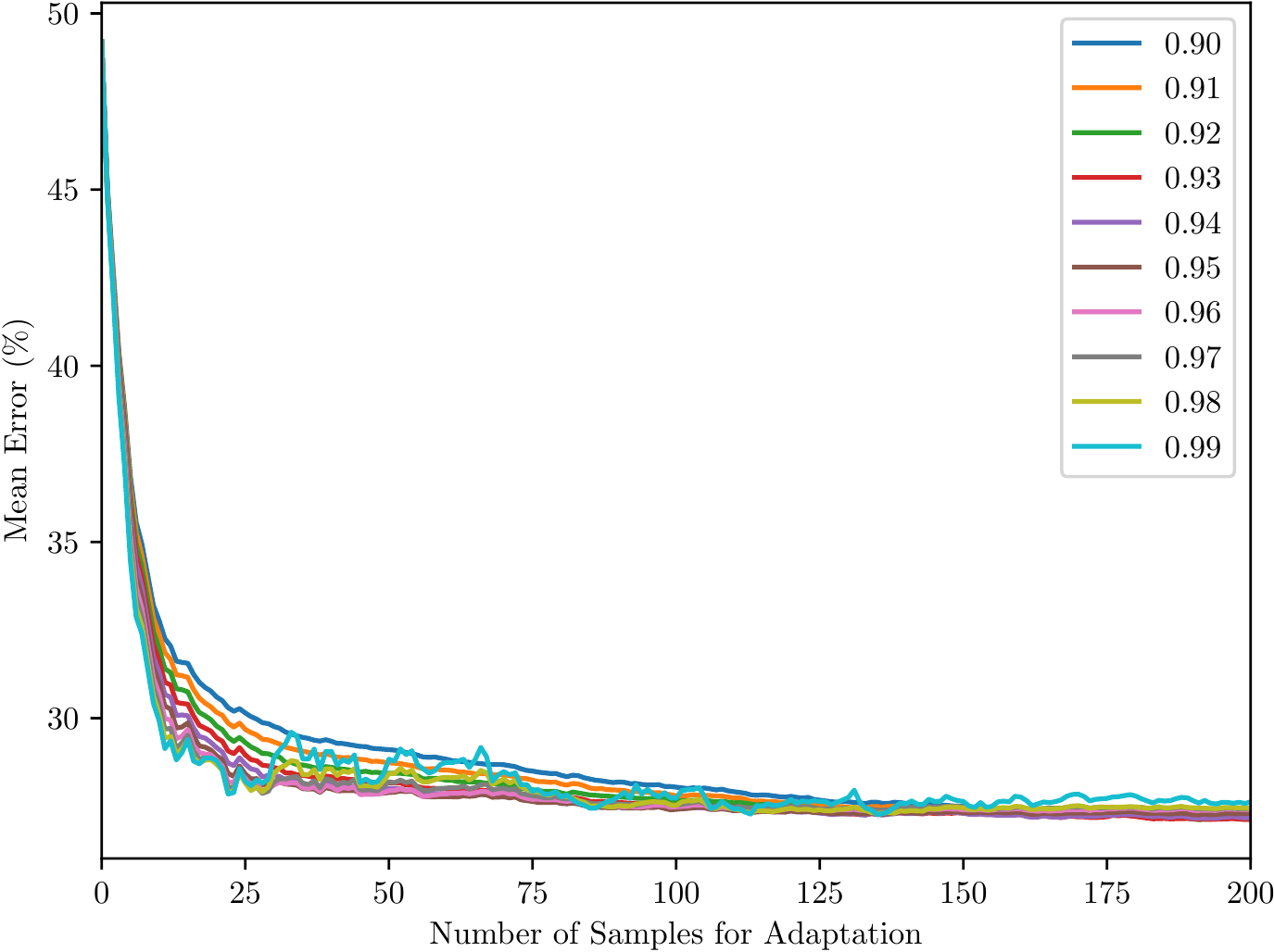}
    \caption{Adaptation results for CIFAR-10C with different values of Momentum Decay Parameter $\omega$. At $\omega = 0.94$, we achieve a good balance between stable adaptation and fast convergence. For all these experiments $\rho_{0}=0.1$ is used. 
    }
    \label{fig:momentum_decay_ablation}
\end{figure}

\subsection{Choice of Momentum Decay Parameter}
 The momentum decay parameter $\omega$ helps to provide stable and fast adaptation on the incoming samples. The effect of varying $\omega$ values is analyzed in Figure~\ref{fig:momentum_decay_ablation}. From this experiment we understand that the adaptation is unstable for a higher value of $\omega$, while it is slower for a lower value. Thus, we use $\omega = 0.94$ for all reported experiments, which empirically provides both stable and fast adaptation. 
\begin{table}
  \centering
  \small
  \begin{tabular}{c|c}
    \toprule
    Corruption Type& Abbreviation\\
    \midrule
    Gaussian Noise &gaus \\
      Shot Noise &shot\\
     Impulse Noise&impul\\
     Defocus Blur&defcs \\
     Glass Blur&gls\\
     Motion Blur&mtn  \\
    Zoom Blur&zm \\
    Snow&snw\\
     Frost&frst  \\
     Fog&fg \\
    Brightness&brt\\
    Contrast&cnt\\
     Elastic&els\\
    Pixelate&px\\
    JPEG Compression&jpg\\
    \bottomrule
    \end{tabular}
    \caption{Abbreviations of corruption types in CIFAR-10/100C and ImageNet-C.}
    \label{tab:abbreviations}
\end{table}
\section{DUA for Additional Adaptation Tasks}
\label{sec:supp-aditional-adaptation-tasks}
\subsection{Domain Adaptation for Digit Recognition}
Table~\ref{tab:Digit_Recog_DA} summarizes results for cross-dataset domain adaptation in digit recognition. For all these experiments we use a simple architecture consisting of 2 convolution layers, 2 batch normalization layers and 2 fully connected layers. Our method improves the results for all the popular domain adaptation benchmarks. Note that we only use random cropping for digit recognition experiments. 

\subsection{Domain Adaptation for Visual Recognition}
In Table~\ref{tab:visda_office}, we list the results obtained by testing on two popular domain adaptation benchmarks,~Office-31~and~VIS-DA. For VIS-DA we take a ResNet-50, pre-trained on ImageNet and then fine tune it on the VIS-DA train split. For the Office-31 dataset, we use an ImageNet pre-trained ResNet-18 which we finetune on the Office-31 train split. 

We show that our method achieves improvements on both Office-31 and VIS-DA datasets. It should be noted that the purpose of these evaluations is not to outperform the state-of-the-art methods which specialize on these tasks. Instead, we show that DUA is applicable to a variety of tasks and architectures. It can be used as an initial adaptation method before applying other established domain adaptation methods such as~\cite{sun2020test, sun2017correlation,ganin2016domain,pinheiro2018unsupervised}.
\subsection{Corruption Benchmarks}
We provide detailed results for the lower severity corruption levels {1--4} (highest severity level 5 is included in the main manuscript) for CIFAR-10C (Table~\ref{tab:cifar_10_results_1_4}), CIFAR-100C (Table~\ref{tab:cifar_100_results_1_4}) and ImageNet-C (Table~\ref{tab:imagenet_c_4_1}).
Note that the abbreviations of the different corruption types are summarized in Table~\ref{tab:abbreviations}.
As can be seen from these results, our DUA achieves consistent improvements over all corruption types and severity levels.

\subsection{Object Detection}
\paragraph{Natural Domain Shifts:} Table~\ref{tab:results-soda10m_day_to_night} shows the results for adapting a detector pre-trained on day images only and tested on night images. Table~\ref{tab:results-kitti_to_soda10m} and Table~\ref{tab:results-soda10m_to_kitti} report the domain adaptation results between KITTI and SODA10M. DUA provides notable gains for all the different scenarios.
\paragraph{Degrading Weather:}We provide results for the lower severity levels of KITTI-Fog and KITTI-Rain~\cite{halder2019physics} in Table~\ref{tab:kitti_fog_rain_supp} (highest severity/lowest visibility is included in the main manuscript). We see that by adapting a model with DUA, the detection performance increases even in lower severities of rain and fog.  

\begin{table*}
  \small
  \centering
  \begin{tabular}{c|cccccc|ccc}
    \toprule
    \multicolumn{1}{c}{Source data:}& SVHN~\cite{netzer2011reading} & SVHN & MNIST~\cite{lecun1998gradient} & MNIST & USPS~\cite{hull1994database} & USPS & \multicolumn{3}{c}{Exemplary Images}\\
    \multicolumn{1}{c}{Target data:}& USPS & MNIST & USPS & SVHN & MNIST & SVHN & MNIST & SVHN & USPS\\
    \midrule
    Source only &66&58&78&16&42&\phantom{0}9&\multirow{3}{*}{\includegraphics[scale = 0.4]{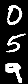}} & \multirow{3}{*}{\includegraphics[scale = 0.4]{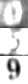}} & \multirow{3}{*}{\includegraphics[scale = 0.4]{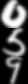}}\\  
    DUA &71&68&86&33&55&28 \\
    Fully Supervised &94&96&95&85&97&91\\
    \bottomrule
  \end{tabular}
  \caption{Domain adaptation results (measured by accuracy in \%) for digit recognition. All results are obtained by using a simple architecture consisting of 2 convolution layers, 2 fully connected layers and batch normalization layers. }
  \label{tab:Digit_Recog_DA}
\end{table*}

\begin{table*}
    \centering
    \small
\subfloat[Office-31~\cite{saenko2010adapting}\label{tab:office-31_results}]{
\begin{tabular}{c|cccccc}
    \toprule
  \multicolumn{1}{c}{Source data:} & Amazon & Amazon & DSLR & DSLR & Webcam & Webcam\\
    \multicolumn{1}{c}{Target data:} & Webcam & DSLR & Webcam & Amazon & DSLR & Amazon\\
    \midrule
    
    Source only &38.4&34.3&12.9&61.4&4.2&66.1\\ 
     DUA&33.2&29.6&\phantom{0}5.7&51.9&2.0&59.7 \\
     Fully Supervised &27.1&11.3&\phantom{0}3.9&34.1&1.2&22.3\\
     \bottomrule
\end{tabular}}\hspace{0.60cm} 
\subfloat[VIS-DA~\cite{peng2017visda}\label{tab:VIS-DA_results}]{
\begin{tabular}{c|c}
        \toprule
            &Classification\\
            &Error (\%) \\
            \midrule
        Source only &57.4\\
        DUA&47.2\\
        Fully Supervised&23.7\\
        \bottomrule
        \end{tabular}}
\caption{Domain adaptation results  (measured by classification error in \%) for a) Office-31 using a ResNet-18 backbone, and for b) VIS-DA using a ResNet-50 backbone.}
\label{tab:visda_office}
\end{table*}

\begin{table*}
  \centering
  \small
  \begin{tabular}{c|ccccccccccccccc|c}
  \toprule
     & gaus & shot & impul & defcs & gls & mtn & zm & snw & frst & fg & brt & cnt & els & px & jpg & mean\\
    \midrule
    \multicolumn{16}{c}{\textbf{Level 4}}\\
    \midrule
    Source &            {63.9} &            {53.7} &            {57.0} &            {28.9} &            {58.9} &            {32.4} &            {38.1} &            {25.9} &            {33.9} &            {17.5} &            {10.4} &            {33.7} &            {26.7} &            {40.7} &            {27.2} &  36.6 \\
    TTT &            {41.5} &            {35.4} &            {39.8} &            {15.0} &            {47.8} & {\bfseries{19.1}} &            {18.4} & {\bfseries{20.1}} &            {24.0} &            {13.5} & {\bfseries{10.0}} & {\bfseries{14.1}} & {\bfseries{17.7}} &            {29.4} & {\bfseries{24.5}} &  24.7 \\
   NORM &            {40.7} &            {37.4} &            {43.2} &            {16.7} &            {47.4} &            {21.8} &            {20.2} &            {29.9} &            {30.3} &            {19.0} &            {16.1} &            {20.5} &            {26.5} &            {26.6} &            {35.7} &  28.8 \\
    DUA & {\bfseries{31.0}} & {\bfseries{27.6}} & {\bfseries{35.8}} & {\bfseries{13.2}} & {\bfseries{40.7}} &            {20.3} & {\bfseries{15.4}} &            {22.2} & {\bfseries{20.6}} & {\bfseries{12.7}} &            {10.1} &            {14.8} &            {20.5} & {\bfseries{18.6}} &            {24.6} &  {\bfseries{21.9}} \\

    \midrule

   Source &            {24.1} &            {17.1} &            {16.4} &            \phantom{0}{6.6} &            {23.5} &            \phantom{0}{8.4} &            \phantom{0}{7.4} &            {12.2} &           {11.5} &            \phantom{0}{8.3} &            \phantom{0}{6.2} &            \phantom{0}{9.2} &           {10.6} &           {19.4} &            {13.1} & 12.9 \\
   
  TENT &            {13.8} & {\bfseries{11.7}} &            {14.3} &            \phantom{0}{6.7} &            {18.6} &            \phantom{0}{8.2} &            \phantom{0}{7.1} &            {10.6} &            \phantom{0}{9.7} &            \phantom{0}{7.5} &            \phantom{0}{6.1} &            \phantom{0}{8.4} &           {10.9} & {\phantom{0}\bfseries{8.5}} &            {13.2} & 10.3 \\
  
   DUA & {\bfseries{13.7}} &            {11.8} & {\bfseries{13.5}} & {\phantom{0}\bfseries{5.9}} & {\bfseries{18.3}} & {\phantom{0}\bfseries{7.6}} & {\phantom{0}\bfseries{6.6}} & {\bfseries{10.3}} & {\phantom{0}\bfseries{9.0}} & {\phantom{0}\bfseries{7.4}} & {\phantom{0}\bfseries{5.8}} & {\phantom{0}\bfseries{7.2}} & {\phantom{0}\bfseries{9.9}} &            \phantom{0}{9.3} & {\bfseries{13.0}} & {\bfseries{10.0}} \\

    \midrule
    \multicolumn{16}{c}{\textbf{Level 3}}\\
    \midrule
   Source & {58.0} &            {47.5} &            {38.5} &            {17.7} &            {46.2} &            {32.8} &            {30.6} &            {22.7} &            {31.8} &            {12.6} &            \phantom{0}{9.5} &            {19.3} &            {20.7} &            {23.7} &            {24.7} &  29.1 \\
    TTT & {37.2} &            {31.6} &            {28.6} &            {11.5} &            {35.8} & {\bfseries{19.1}} &            {15.8} & {\bfseries{17.8}} &            {23.3} & {\bfseries{11.0}} & {\phantom{0}\bfseries{9.1}} & {\bfseries{11.6}} & {\bfseries{14.3}} &            {18.9} & {\bfseries{22.3}} &  20.5 \\
   NORM & {37.8} &            {35.1} &            {34.7} &            {14.1} &            {38.2} &            {21.7} &            {18.2} &            {27.5} &            {29.0} &            {16.6} &           {15.2} &            {18.6} &            {19.6} &            {21.1} &            {33.3} &  25.4 \\
    DUA & {28.3} & {\bfseries{24.6}} & {\bfseries{27.0}} & {\bfseries{10.4}} & {\bfseries{30.7}} &            {20.2} & {\bfseries{14.4}} &            {20.4} & {\bfseries{19.3}} & {\bfseries{11.0}} &            \phantom{0}{9.2} &            {12.3} &            {14.6} & {\bfseries{15.1}} &            {23.1} &  \bfseries{18.7} \\
    \midrule
    Source &            {20.4} &            {14.6} &            \phantom{0}{9.7} & {\phantom{0}\bfseries{5.4}} &            {12.9} &           \phantom{0}{8.6} &            \phantom{0}{6.5} &            \phantom{0}{9.9} &           {11.4} &            \phantom{0}{6.3} &            \phantom{0}{5.5} &            \phantom{0}{7.2} &            \phantom{0}{7.4} &            \phantom{0}{9.6} &            {12.1} &  \phantom{0}9.8 \\
  TENT &            {12.6} & {\bfseries{10.4}} &           {10.4} &            \phantom{0}{6.0} &            {12.7} &            \phantom{0}{8.1} &            \phantom{0}{6.7} &            \phantom{0}{9.5} & {\phantom{0}\bfseries{9.1}} &            \phantom{0}{6.7} &            \phantom{0}{5.9} &            \phantom{0}{7.5} &            \phantom{0}{8.4} &            \phantom{0}{7.5} &            {12.7} &  \phantom{0}8.9 \\
   DUA & {\bfseries{12.2}} &            {10.5} & {\phantom{0}\bfseries{9.3}} &            \phantom{0}{5.5} & {\bfseries{11.9}} & {\phantom{0}\bfseries{7.8}} & {\phantom{0}\bfseries{6.1}} & {\phantom{0}\bfseries{9.1}} & {\phantom{0}\bfseries{9.1}} & {\phantom{0}\bfseries{6.1}} & {\phantom{0}\bfseries{5.4}} & {\phantom{0}\bfseries{6.3}} & {\phantom{0}\bfseries{7.0}} & {\phantom{0}\bfseries{7.4}} & {\bfseries{11.7}} &  \phantom{0}\bfseries{8.4} \\

    \midrule
    \multicolumn{16}{c}{\textbf{Level 2}}\\
    \midrule
    Source &            {43.1} &            {27.8} &            {29.3} &           {10.2} &            {49.5} &            {23.4} &            {22.4} &            {26.4} &            {21.3} &           {10.3} &            \phantom{0}{8.7} &            {13.4} &            {14.7} &            {17.9} &            {22.3} &  22.7 \\
    TTT &            {28.8} &            {20.7} &            {23.0} & \phantom{0}{\bfseries{9.0}} &            {36.6} & {\bfseries{15.4}} &            {13.1} & {\bfseries{20.2}} &            {16.9} & {\phantom{0}\bfseries{9.2}} & {\phantom{0}\bfseries{8.3}} & {\bfseries{10.2}} & {\bfseries{12.5}} &            {14.8} & {\bfseries{19.7}} &  17.2 \\
   NORM &            {31.0} &            {25.3} &            {28.7} &           {13.5} &            {38.8} &            {18.8} &            {16.3} &            {27.8} &            {23.9} &           {15.4} &           {14.6} &            {17.1} &            {18.7} &            {19.6} &            {30.6} &  22.7 \\
    DUA & {\bfseries{22.3}} & {\bfseries{16.8}} & {\bfseries{22.9}} &            \phantom{0}{9.2} & {\bfseries{30.3}} &            {16.0} & {\bfseries{12.7}} &            {21.5} & {\bfseries{15.7}} &            \phantom{0}{9.6} &            \phantom{0}{8.7} &            {11.1} &            {12.7} & {\bfseries{13.3}} &            {20.8} &  \bfseries{16.2} \\

    \midrule
   Source &            {13.4} &            \phantom{0}{8.8} &            \phantom{0}{8.0} & \phantom{0}{\bfseries{5.1}} &            {14.2} &            \phantom{0}{6.5} &            \phantom{0}{5.8} &            \phantom{0}{9.2} &            \phantom{0}{8.5} &            \phantom{0}{5.3} &            \phantom{0}{5.3} &            \phantom{0}{6.1} &            \phantom{0}{6.5} &            \phantom{0}{7.8} & {\bfseries{10.9}} &  \phantom{0}8.1 \\
  TENT &            {10.2} &            \phantom{0}{7.6} &            \phantom{0}{8.6} &            \phantom{0}{5.9} &            {13.0} &            \phantom{0}{7.2} &            \phantom{0}{6.2} & {\phantom{0}\bfseries{8.1}} &            \phantom{0}{7.8} &            \phantom{0}{6.3} &            \phantom{0}{5.8} &            \phantom{0}{6.9} &            \phantom{0}{7.5} &            \phantom{0}{7.0} &            {11.8} &  \phantom{0}8.0 \\
   DUA & {\bfseries{10.0}} & {\phantom{0}\bfseries{7.5}} & {\phantom{0}\bfseries{7.6}} & {\phantom{0}\bfseries{5.1}} & {\bfseries{12.4}} & {\phantom{0}\bfseries{6.4}} & {\phantom{0}\bfseries{5.7}} &            \phantom{0}{8.3} & \phantom{0}{\bfseries{7.3}} & \phantom{0}{\bfseries{5.2}} & \phantom{0}{\bfseries{5.2}} & \phantom{0}{\bfseries{5.7}} & \phantom{0}{\bfseries{6.4}} & \phantom{0}{\bfseries{6.8}} & {\bfseries{10.9}} &  \phantom{0}\bfseries{7.4} \\

    \midrule
    \multicolumn{16}{c}{\textbf{Level 1}}\\
    \midrule
    Source &            {25.8} &            {18.4} &            {19.0} &            \phantom{0}{8.5} &            {51.1} &            {14.7} &            {18.2} &            {15.0} &            {13.8} &            \phantom{0}{8.3} &            \phantom{0}{8.3} &            \phantom{0}{8.7} &            {14.4} &            {11.3} &            {16.5} &  16.8 \\
    TTT &            {19.1} &            {15.8} & {\bfseries{16.5}} & \phantom{0}{\bfseries{8.0}} &            {37.9} & {\bfseries{11.7}} & {\bfseries{12.2}} & {\bfseries{12.8}} & {\bfseries{11.9}} & \phantom{0}{\bfseries{8.2}} & \phantom{0}{\bfseries{8.0}} & \phantom{0}{\bfseries{8.3}} & {\bfseries{12.6}} &            {11.1} & {\bfseries{15.5}} &  14.0 \\
   NORM &            {24.0} &            {20.9} &            {22.5} &           {13.4} &            {38.1} &            {16.5} &            {15.5} &            {20.5} &            {18.8} &           {14.9} &           {14.0} &           {15.3} &            {19.1} &            {16.9} &            {24.7} &  19.7 \\
    DUA & {\bfseries{16.5}} & {\bfseries{13.8}} &            {16.6} &            \phantom{0}{8.3} & {\bfseries{30.4}} &            {12.4} &            {12.6} &            {14.5} &            {12.2} &            \phantom{0}{8.4} &            \phantom{0}{8.4} &            \phantom{0}{8.8} &            {13.4} & {\bfseries{11.0}} &            {15.9} &  \bfseries{13.6} \\
    \midrule
    Source &            \phantom{0}{8.7} &            \phantom{0}{6.5} & \phantom{0}{\bfseries{6.2}} & \phantom{0}{\bfseries{4.9}} &            {14.1} & \phantom{0}{\bfseries{5.5}} &            \phantom{0}{5.9} &            \phantom{0}{6.4} &            \phantom{0}{6.5} & \phantom{0}{\bfseries{4.9}} & \phantom{0}{\bfseries{5.0}} & \phantom{0}{\bfseries{5.0}} & \phantom{0}{\bfseries{6.9}} & \phantom{0}{\bfseries{5.8}} &            \phantom{0}{8.7} &  \phantom{0}6.7 \\
  TENT &            \phantom{0}{7.5} &            \phantom{0}{6.9} &            \phantom{0}{7.2} &            \phantom{0}{5.7} &            {12.4} &            \phantom{0}{6.2} &            \phantom{0}{6.3} &            \phantom{0}{6.8} &            \phantom{0}{6.5} &            \phantom{0}{5.9} &            \phantom{0}{5.7} &            \phantom{0}{6.0} &            \phantom{0}{7.9} &            \phantom{0}{6.5} &            \phantom{0}{9.1} &  \phantom{0}7.1 \\
   DUA & \phantom{0}{\bfseries{7.3}} & \phantom{0}{\bfseries{6.2}} & \phantom{0}{\bfseries{6.2}} &            \phantom{0}{5.1} & {\bfseries{11.9}} & \phantom{0}{\bfseries{5.5}} & \phantom{0}{\bfseries{5.8}} & \phantom{0}{\bfseries{6.2}} & \phantom{0}{\bfseries{6.1}} &            \phantom{0}{5.1} &            \phantom{0}{5.1} &            \phantom{0}{5.1} &            \phantom{0}{7.0} & \phantom{0}{\bfseries{5.8}} & \phantom{0}{\bfseries{8.5}} &  \phantom{0}\bfseries{6.5} \\

    \bottomrule
  \end{tabular}
  \caption{Error (\%) for each corruption in CIFAR-10C severity (Level 1--4) is reported. Source refers to results obtained from a model trained on clean train set and tested on corrupted test sets. For a fair comparison with TTT and NORM, ResNet-26 is used. For comparison with TENT we take the Wide-ResNet-40-2 model from their official Github Repository. Lowest error is highlighted for each corruption.}
  \label{tab:cifar_10_results_1_4}
\end{table*}

\begin{table*}
  \centering
  \small
  \begin{tabular}{c|ccccccccccccccc|c}
  \toprule
     & gaus & shot & impul & defcs & gls & mtn & zm & snw & frst & fg & brt & cnt & els & px & jpg & mean\\
    \midrule
    \multicolumn{16}{c}{\textbf{Level 4}}\\
    \midrule
    Source &            {88.3} &            {85.3} &            {92.8} &            {54.3} &            {84.8} &            {57.3} &            {57.2} &            {54.9} &            {65.9} &            {48.6} & {\bfseries{35.8}} &            {61.1} &            {52.9} &            {73.0} &            {61.0} & 64.9 \\
   TTT &            {81.6} &            {78.3} &            {81.1} &            {48.6} &            {78.7} &            {52.5} &            {53.4} &            {53.8} &            {62.8} &            {49.5} &            {38.5} &            {50.3} &            {50.2} &            {66.2} & {\bfseries{57.2}} & 60.2 \\
  NORM &            {70.5} &            {67.6} &            {72.1} & {\bfseries{41.1}} &            {69.9} & {\bfseries{46.1}} &            {44.5} &            {54.7} &            {55.2} &            {46.7} &            {38.8} & {\bfseries{44.1}} &            {50.8} &            {49.9} &            {64.2} & 54.4 \\
   DUA & {\bfseries{66.0}} & {\bfseries{62.9}} & {\bfseries{66.4}} &            {41.3} & {\bfseries{66.5}} &            {48.2} & {\bfseries{43.0}} & {\bfseries{53.5}} & {\bfseries{52.8}} & {\bfseries{43.2}} &            {37.6} &            {45.7} & {\bfseries{48.2}} & {\bfseries{46.6}} &            {57.6} & \bfseries{52.0} \\

    \midrule
    Source &            {60.7} &            {51.6} &            {47.9} &            {27.1} &            {54.4} &            {30.3} &            {28.9} &            {37.4} &            {39.0} &            {35.4} &            {27.2} &            {35.9} &            {34.4} &            {39.0} &            {40.1} & 39.3 \\
  TENT & {\bfseries{38.9}} & {\bfseries{36.3}} & {\bfseries{36.6}} &            {27.3} & {\bfseries{42.0}} & {\bfseries{28.9}} &            {28.4} & {\bfseries{34.8}} & {\bfseries{32.8}} & {\bfseries{32.1}} & {\bfseries{26.3}} & {\bfseries{29.8}} & {\bfseries{33.3}} & {\bfseries{29.9}} & {\bfseries{38.4}} & \bfseries{33.1} \\
   DUA &            {43.0} &            {39.0} &            {37.9} & {\bfseries{26.9}} &            {44.7} &            {29.5} & {\bfseries{28.3}} &            {36.5} &            {34.3} &            {33.8} &            {26.6} &            {31.0} &            {33.8} &            {31.0} &            {38.9} & 34.3 \\
    
    \midrule
    \multicolumn{16}{c}{\textbf{Level 3}}\\
    \midrule
    Source &            {86.0} &            {81.4} &            {83.8} &            {42.7} &            {78.3} &            {57.7} &            {52.4} &            {53.0} &            {64.1} &            {40.1} & {\bfseries{33.3}} &            {49.3} &            {46.6} &            {54.4} &            {58.2} & 58.8 \\
   TTT &            {79.6} &            {74.6} &            {69.3} &            {42.5} &            {73.0} &            {53.2} &            {49.8} &            {51.2} &            {61.4} &            {42.1} &            {36.8} &            {43.5} &            {45.8} &            {52.9} & {\bfseries{55.2}} & 55.4 \\
  NORM &            {67.7} &            {64.6} &            {62.3} &            {60.9} &            {63.4} & {\bfseries{42.6}} &            {52.6} &            {54.5} & {\bfseries{41.5}} & {\bfseries{37.8}} &            {37.8} & {\bfseries{41.1}} &            {44.3} &            {45.0} &            {61.7} & 51.9 \\
   DUA & {\bfseries{63.8}} & {\bfseries{60.0}} & {\bfseries{57.9}} & {\bfseries{37.3}} & {\bfseries{58.4}} &            {48.2} & {\bfseries{41.2}} & {\bfseries{50.8}} &            {52.4} &            {39.2} &            {35.5} &            {41.4} & {\bfseries{41.7}} & {\bfseries{42.1}} &            {55.3} & \bfseries{48.4} \\
    \midrule
    Source &            {55.2} &            {45.9} &            {36.9} &            {25.7} &            {39.9} &            {30.5} &            {27.4} &            {33.3} &            {38.1} &            {29.5} &            {25.5} &            {30.5} &            {28.6} &            {30.3} &            {38.0} & 34.4 \\
  TENT & {\bfseries{37.1}} & {\bfseries{34.3}} & {\bfseries{32.0}} &            {26.1} & {\bfseries{34.6}} & {\bfseries{29.2}} &            {27.7} & {\bfseries{32.3}} & {\bfseries{32.3}} & {\bfseries{29.0}} &            {25.6} & {\bfseries{28.4}} &            {29.2} & {\bfseries{28.3}} &            {37.4} & \bfseries{30.9} \\
   DUA &            {40.8} &            {36.5} &            {32.2} & {\bfseries{25.3}} &            {37.0} &            {29.6} & {\bfseries{27.1}} &            {32.7} &            {34.4} & {\bfseries{29.0}} & {\bfseries{25.2}} &            {28.6} & {\bfseries{28.4}} &            {28.7} & {\bfseries{37.3}} & 31.5 \\
    \midrule
    \multicolumn{16}{c}{\textbf{Level 2}}\\
    \midrule
    Source &            {79.9} &            {66.1} &            {73.7} & {\bfseries{33.7}} &            {79.6} &            {48.9} &            {46.4} &            {56.8} &            {52.4} & {\bfseries{35.2}} & {\bfseries{31.3}} &            {42.1} &            {40.5} &            {48.0} &            {55.2} & 52.7 \\
   TTT &            {71.5} &            {61.8} &            {59.8} &            {36.5} &            {73.1} &            {47.2} &            {46.0} &            {55.7} &            {52.8} &            {38.0} &            {35.2} &            {39.7} &            {42.0} &            {47.9} & {\bfseries{52.6}} & 50.7 \\
  NORM &            {61.1} &            {54.9} &            {55.9} &            {36.3} &            {61.1} & {\bfseries{43.0}} &            {40.7} &            {53.0} &            {49.0} &            {39.6} &            {37.0} &            {39.5} &            {42.8} &            {43.3} &            {59.0} & 47.7 \\
   DUA & {\bfseries{57.1}} & {\bfseries{50.7}} & {\bfseries{52.0}} &            {35.2} & {\bfseries{58.4}} &            {43.8} & {\bfseries{39.1}} & {\bfseries{52.8}} & {\bfseries{47.6}} &            {35.9} &            {34.2} & {\bfseries{38.9}} & {\bfseries{39.9}} & {\bfseries{39.7}} &            {52.9} & \bfseries{45.2} \\

    \midrule
    Source &            {44.6} &            {34.5} &            {30.7} &            {24.3} &            {41.5} &            {27.7} &            {26.2} &            {32.7} &            {31.8} &            {26.8} &            {24.4} &            {27.5} & {\bfseries{27.9}} &            {28.0} &            {36.5} & 31.0 \\
  TENT & {\bfseries{33.3}} & {\bfseries{29.5}} & {\bfseries{28.8}} &            {25.7} & {\bfseries{34.9}} &            {27.4} &            {27.0} & {\bfseries{30.6}} & {\bfseries{29.6}} &            {27.0} &            {25.4} &            {27.3} &            {29.1} & {\bfseries{27.6}} &            {36.3} & \bfseries{29.3} \\
   DUA &            {35.9} &            {31.2} & {\bfseries{28.8}} & {\bfseries{24.1}} &            {36.9} & {\bfseries{27.2}} & {\bfseries{26.0}} &            {32.2} &            {30.9} & {\bfseries{26.1}} & {\bfseries{24.0}} & {\bfseries{26.6}} & {\bfseries{27.9}} & {\bfseries{27.6}} & {\bfseries{35.9}} & 29.4 \\

    \midrule
    \multicolumn{16}{c}{\textbf{Level 1}}\\
    \midrule
    Source &            {65.0} &            {53.4} &            {54.2} & {\bfseries{30.3}} &            {80.1} &            {40.2} &            {42.9} & {\bfseries{39.6}} &            {43.0} & {\bfseries{31.0}} & {\bfseries{30.2}} & {\bfseries{31.8}} & {\bfseries{40.3}} &            {37.0} &            {48.1} & 44.5 \\
   TTT &            {60.4} &            {53.0} &            {48.0} &            {34.7} &            {74.0} &            {41.3} &            {41.3} &            {41.5} &            {44.2} &            {34.6} &            {34.4} &            {34.8} &            {41.8} &            {39.4} & {\bfseries{47.0}} & 44.7 \\
  NORM &            {52.9} &            {48.8} &            {47.7} &            {36.2} &            {60.6} &            {40.1} &            {39.5} &            {43.9} &            {44.0} &            {36.8} &            {36.3} &            {36.7} &            {43.2} &            {40.6} &            {52.0} & 44.0 \\
   DUA & {\bfseries{49.6}} & {\bfseries{45.5}} & {\bfseries{44.4}} &            {32.5} & {\bfseries{58.0}} & {\bfseries{39.4}} & {\bfseries{38.3}} &            {41.7} & {\bfseries{41.9}} &            {33.2} &            {33.1} &            {33.5} &            {40.8} & {\bfseries{36.0}} & {\bfseries{47.0}} & \bfseries{41.0} \\
    \midrule
    Source &            {34.4} &            {29.6} &            {26.9} & {\bfseries{23.8}} &            {42.9} &            {25.6} &            {26.1} & {\bfseries{26.1}} &            {27.4} & {\bfseries{24.0}} & {\bfseries{23.8}} &            {24.3} & {\bfseries{28.4}} & {\bfseries{25.2}} &            {32.4} & 28.1 \\
  TENT & {\bfseries{29.3}} & {\bfseries{27.6}} &            {26.9} &            {25.5} & {\bfseries{34.7}} &            {26.6} &            {26.7} &            {27.0} &            {27.3} &            {25.6} &            {25.2} &            {26.0} &            {29.8} &            {26.7} &            {32.9} & 27.9 \\
   DUA &            {31.2} &            {28.5} & {\bfseries{26.4}} & {\bfseries{23.8}} &            {36.9} & {\bfseries{25.3}} & {\bfseries{25.9}} &            {26.2} & {\bfseries{27.1}} &            {24.1} &            {23.9} & {\bfseries{23.9}} &            {28.6} &            {25.4} & {\bfseries{32.0}} & \bfseries{27.3} \\

    \bottomrule
  \end{tabular}
  \caption{Error (\%) for each corruption in CIFAR-100C severity (Level 1--4) is reported. Source refers to results obtained from a model trained on clean train set and tested on corrupted test sets. For a fair comparison with TTT and NORM, ResNet-26 is used. For comparison with TENT we take the Wide-ResNet-40-2 model from their official Github Repository. Lowest error is highlighted for each corruption.}
  \label{tab:cifar_100_results_1_4}
\end{table*}

\begin{table*}
  \centering
  \small
  \begin{tabular}{c|ccccccccccccccc|c}
  \toprule
     & gaus & shot & impul & defcs & gls & mtn & zm & snw & frst & fg & brt & cnt & els & px & jpg & mean\\
    \midrule
    \multicolumn{16}{c}{\textbf{Level 4}}\\
    \midrule
    Source &            {93.2} &            {94.7} &            {94.3} &            {84.5} &            {89.4} &            {85.3} &            {77.2} &            {83.4} &            {79.4} &            {72.8} &            {44.5} &            {88.1} &            {63.4} &            {71.2} &            {58.8} & 78.7 \\
  NORM &            {84.4} & {\bfseries{77.6}} &            {87.3} &            {86.4} &            {88.0} & {\bfseries{76.8}} &            {70.7} &            {76.9} & {\bfseries{70.9}} & {\bfseries{56.0}} & {\bfseries{37.8}} & {\bfseries{64.4}} & {\bfseries{53.5}} & {\bfseries{58.6}} &            {57.8} & \bfseries{69.8} \\
   DUA & {\bfseries{78.1}} &            {82.8} & {\bfseries{80.3}} & {\bfseries{82.8}} & {\bfseries{83.3}} &            {78.7} & {\bfseries{69.8}} & {\bfseries{76.1}} &            {74.2} &            {59.7} &            {40.4} &            {87.3} &            {55.8} &            {61.8} & {\bfseries{54.9}} & 71.1 \\

    \midrule
    \multicolumn{16}{c}{\textbf{Level 3}}\\
    \midrule
    Source &            {80.9} &            {82.7} &            {82.9} &            {74.1} &            {85.4} &            {73.9} &            {71.9} &            {73.6} &            {77.8} &            {66.2} &            {39.6} &            {65.3} &            {51.1} &            {56.7} &            {49.3} & 68.8 \\
  NORM & {\bfseries{65.5}} &            {69.8} & {\bfseries{63.2}} &            {71.1} &            {79.5} & {\bfseries{63.6}} & {\bfseries{58.2}} &            {68.1} & {\bfseries{65.4}} &            {55.3} &            {39.3} & {\bfseries{55.0}} & {\bfseries{41.0}} &            {51.8} &            {50.9} & \bfseries{59.8} \\
   DUA &            {67.6} & {\bfseries{68.5}} &            {69.0} & {\bfseries{70.7}} & {\bfseries{78.6}} &            {66.4} &            {65.3} & {\bfseries{66.9}} &            {70.6} & {\bfseries{54.3}} & {\bfseries{37.4}} &            {61.8} &            {46.5} & {\bfseries{47.6}} & {\bfseries{48.2}} & 61.3 \\

    \midrule
    \multicolumn{16}{c}{\textbf{Level 2}}\\
    \midrule
    Source &            {63.6} &            {67.8} &            {74.6} &            {59.4} &            {65.4} &            {57.7} &            {65.2} &            {77.5} &            {67.0} &            {56.3} &            {36.4} &            {51.5} &            {60.9} &            {43.3} &            {46.3} & 59.5 \\
  NORM & {\bfseries{50.7}} &            {61.2} &            {65.2} &            {61.3} & {\bfseries{54.4}} & {\bfseries{46.4}} & {\bfseries{55.8}} & {\bfseries{69.1}} & {\bfseries{58.0}} & {\bfseries{45.6}} &            {36.1} & {\bfseries{40.3}} &            {64.1} &            {41.3} & {\bfseries{40.6}} & \bfseries{52.7} \\
   DUA &            {56.4} & {\bfseries{58.0}} & {\bfseries{62.0}} & {\bfseries{56.5}} &            {62.0} &            {53.1} &            {58.0} &            {69.9} &            {63.0} &            {49.3} & {\bfseries{35.8}} &            {50.5} & {\bfseries{56.6}} & {\bfseries{40.6}} &            {45.3} & 54.5 \\

    \midrule
    \multicolumn{16}{c}{\textbf{Level 1}}\\
    \midrule
   Source &            {50.5} &            {53.1} &            {61.9} &            {51.3} &            {52.4} &            {45.2} &            {55.9} &            {55.5} &            {49.7} &            {49.0} &            {34.2} &            {44.0} &            {40.4} &            {41.3} &            {42.6} & 48.5 \\
  NORM &            {47.0} & {\bfseries{48.4}} & {\bfseries{54.1}} & {\bfseries{50.2}} & {\bfseries{47.8}} & {\bfseries{39.5}} &            {50.3} & {\bfseries{47.7}} & {\bfseries{44.1}} & {\bfseries{41.1}} &            {38.4} & {\bfseries{36.6}} &            {42.0} & {\bfseries{37.0}} &            {43.2} & \bfseries{44.5} \\
   DUA & {\bfseries{45.7}} &            {49.0} &            {54.7} &            {50.6} &            {49.8} &            {43.4} & {\bfseries{50.1}} &            {51.7} &            {46.9} &            {45.8} & {\bfseries{31.8}} &            {43.3} & {\bfseries{38.9}} &            {39.1} & {\bfseries{41.5}} & 45.5 \\
    
    \bottomrule
  \end{tabular}
  \caption{Error (\%) for each corruption in ImageNet-C severity (Level 1--4) is reported.  Source refers to results obtained from a model pre-trained on ImageNet and tested on corrupted test sets. All results are obtained by using ResNet-18 as the backbone architecture. Lowest error is highlighted for each corruption.}
  \label{tab:imagenet_c_4_1}
\end{table*}

\begin{table*}
    \centering
    \small
    \subfloat[SODA10M~\cite{han2021soda10m} (Day $\rightarrow$ Night)\label{tab:results-soda10m_day_to_night}]{
\begin{tabular}{c|ccc}
    \toprule
     & Car & Ped & Cyclist \\
    \midrule
    SO & 75.3 &48.3 &50.1\\
    DUA & 77.2 &49.9 &51.6\\
    FS & 86.9 &55.7 &63.2\\
    \bottomrule
  \end{tabular}}\hspace{1cm}
  \subfloat[KITTI~\cite{geiger2013vision} $\rightarrow$ SODA10M\label{tab:results-kitti_to_soda10m}]{
\begin{tabular}{c|ccc}
    \toprule
     & Car & Ped & Cyclist \\
    \midrule
    SO & 43.6 &13.9 &18.9\\
    DUA & 49.4 &17.2 &23.1\\
    FS & 84.8 &46.1 &61.2\\
    \bottomrule
  \end{tabular}}\hspace{1cm}
  \subfloat[SODA10M $\rightarrow$ KITTI \label{tab:results-soda10m_to_kitti}]{
\begin{tabular}{c|ccc}
    \toprule
     & Car & Ped & Cyclist \\
    \midrule
    SO & 80.1 &55.5 &33.9\\
    DUA & 81.3 &56.3 &35.1\\
    FS & 91.8 &71.3 &76.5\\
    \bottomrule
\end{tabular}}
\caption{Domain adaptation results (mAP@50) for object detection with YOLOv3. a) We train on the SODA10M day split and test on the SODA10M night split. b) We train on the KITTI train split and test on the SODA10M test split. c) We train on the SODA10M day split and test on the KITTI validation split (details in Table~\ref{tab:num_samples}). SO: source-only, FS: fully supervised.}
\label{tab:kitti_soda_results}
\end{table*}

\begin{table*}
    \centering
    \small
 \subfloat[KITTI $\rightarrow$ KITTI-Fog~\cite{halder2019physics}\label{tab:kitti_fog_rain:fog:sup}]{
\begin{tabular}{c|ccc}
    \toprule
     & Car & Pedestrian & Cyclist \\
     \midrule
    \multicolumn{4}{c}{40m fog visibility}\\
    \midrule
    Source only & 39.6 &39.3 &20.2\\
    DUA & 61.3 &53.5 &39.9\\
    Fully Supervised & 77.8 &68.1 &70.5\\
    \midrule
    \multicolumn{4}{c}{50m fog visibility}\\
    \midrule
    Source only & 48.2&45.9&27.3\\
    DUA & 69.6&59.7&48.4\\
    Fully Supervised & 82.2 &71.3 &74.2\\
    \bottomrule
  \end{tabular}}\hspace{0.5cm}
   \subfloat[KITTI $\rightarrow$ KITTI-Rain~\cite{halder2019physics}\label{tab:kitti_fog_rain:rain:sup}]{
\begin{tabular}{c|ccc}
  \toprule
   & Car & Pedestrian & Cyclist \\
  \midrule
    \multicolumn{4}{c}{100 mm/hr rain}\\
    \midrule
    Source only & 93.2&80.5&81.1\\
    DUA & 94.9&81.7&83.2\\
    Fully Supervised & 95.4 &83.2 &84.1\\
    
    \midrule
    \multicolumn{4}{c}{75 mm/hr rain}\\
    \midrule
    Source only & 95.5 &82.6 &83.6\\
    DUA & 96.1 &83.2 &86.0\\
    Fully Supervised & 96.6 &84.1 &87.2\\
    \bottomrule
\end{tabular}}
\caption{Results for KITTI pretrained YOLOv3 tested on rain and fog datasets. Mean Average Precision (mAP@50) is reported for the three common classes in KITTI dataset. a) Results for $40$m and $50$m visibility in fog. b) Results for $100$mm/hr and $75$mm/hr rain intensity.}
\label{tab:kitti_fog_rain_supp}
\end{table*}

\end{appendices}
\end{document}